\documentclass{article}

\usepackage{PRIMEarxiv}

\usepackage[utf8]{inputenc} % allow utf-8 input
\usepackage[T1]{fontenc}    % use 8-bit T1 fonts
\usepackage{hyperref}       % hyperlinks
\usepackage{url}            % simple URL typesetting
\usepackage{booktabs}       % professional-quality tables
\usepackage{amsfonts}       % blackboard math symbols
\usepackage{nicefrac}       % compact symbols for 1/2, etc.
\usepackage{microtype}      % microtypography
\usepackage{lipsum}
\usepackage{multirow}
\usepackage{wrapfig}
\usepackage{subcaption}
\usepackage{amssymb}
\usepackage{amsmath,amsfonts,bm}
\usepackage{mathrsfs}
\usepackage{natbib}  % DO NOT CHANGE THIS AND DO NOT ADD ANY OPTIONS TO IT
\usepackage{caption} % DO NOT CHANGE THIS AND DO NOT ADD ANY OPTIONS TO IT
\usepackage{fancyhdr}       % header
\usepackage{graphicx}       % graphics
\graphicspath{{media/}}     % organize your images and other figures under media/ folder

\title{Based on What We Can Control Artificial Neural Networks
}

\author{
  Cheng~Kang \\
  Czech Technical University \\
  Prague, Czech Republic\\
  \texttt{kangchen@fel.cvut.cz} \\
   \And
  Xujing~Yao \\
  University of Leicester \\
  Leicester, UK\\
  \texttt{xy147@le.ac.uk} \\
}

\begin{document}
\maketitle

\begin{abstract}
How can the stability and efficiency of Artificial Neural Networks (ANNs) be ensured through a systematic analysis method? This paper seeks to address that query. While numerous factors can influence the learning process of ANNs, utilizing knowledge from control systems allows us to analyze its system function and simulate system responses. Although the complexity of most ANNs is extremely high, we still can analyze each factor (e.g., optimiser, hyperparameters) by simulating their system response. This new method also can potentially benefit the development of new optimiser and learning system, especially when discerning which components adversely affect ANNs. Controlling ANNs can benefit from the design of optimiser and learning system, as (1) all optimisers act as controllers, (2) all learning systems operate as control systems with inputs and outputs, and (3) the optimiser should match the learning system. Please find codes: \url{https://github.com/RandomUserName2023/Control-ANNs}.  
\end{abstract}

% keywords can be removed
\keywords{Optimizer \and Controller \and Learning System \and Control System \and Fuzzy Logic \and Filter}

\section{Introduction}

Controlling artificial neural networks (ANNs) has become an urgent issue on such a dramatically growing domain. Although ANN models, such as, vision models (e.g., CNN \cite{krizhevsky2012imagenet}, VGG19 \cite{simonyan2014very}, ResNet50 \cite{he2016deep}, EfficientNet \cite{tan2019efficientnet}, ViT \cite{dosovitskiy2020image}), language models (e.g., BERT \cite{devlin2018bert}, GPT \cite{radford2018improving}, PaLM \cite{chowdhery2022palm}), and generative models (e.g., GAN \cite{2014Generative}, VAE \cite{kingma2013auto}, Stable Diffusion Models \cite{ho2020denoising,rombach2022high}), all require input and output, as they aim to map the gap between their output and the desired output. However, basically, CNN-based vision models prefer SGDM \cite{qian1999momentum} optimiser, and generative models tend to rely on AdaM optimiser. Using various architecture on CNN-based vision models (e.g., from VGG19 to ResNet50, from GAN to CycleGAN \cite{zhu2017unpaired}, and from CNN to FFNN \cite{hinton2022forward}) yield significantly varied results for classification and generation tasks. Two critical questions arise: \textbf{(1)} why some of them satisfy the corresponding optimiser, \textbf{(2)} based on what to propose an advanced ANN architecture and a proper optimiser.

Compared to existing era-acrossing optimisers, such as SGD \cite{robbins1951stochastic,cotter2011better,zhou2017convergence}, SGDM \cite{qian1999momentum,liu2020improved}, AdaM \cite{kingma2014adam,bock2018improvement}, PID \cite{wang2020pid}, and Gaussian LPF-SGD \cite{bisla2022low}, we proposed a FuzzyPID optimiser modified by fuzzy logic to avoid vibration during PID optimiser learning process. Referring to Gaussian LPF-SGD (GLFP-SGD), we also proposed two filter processed SGD methods according to the low and high frequency part during the SGD optimiser learning process: low-pass-filter SGD (LPF-SGD) and high-pass-filter SGD (HPF-SGD). To achieve stable and convergent performance, we simulate these above optimisers on the system response to analyze their attributes. When using simple and straightforward architecture (without high techniques, such as, BN \cite{ioffe2015batch}, ReLU \cite{nair2010rectified}, pooling \cite{wu2015max}, and exponential or cosine decay \cite{li2021second}), we found their one step system response are always consistent with their training process. Therefore, we conclude that every optimiser actually can be considered as a controller that optimise the training process. Results using HPF-SGD indicate that the high frequency part using SGD optimiser significantly benefits the learning process and the classification performance.

To analyze the learning progress of most ANNs, for example, CNN using backpropagation algorithm, FFNN using forward-forward algorithm, and GAN such a generative model using random noise to generate samples. We assume above three mentioned models here essentially can be represented by corresponding control systems. But the difficulty is that when using different optimisers, especially, AdaM, we cannot analyze its stability and convergence, as the complexity is extremely high. Thus, we use MATLAB Simulink to analyze their system response, as well as their generating response. Experiment results indicate that advanced architectures and designs of these three ANNs can improve the learning, such as residual connections (RSs) on ResNets, a higher Threshold on FFNN, and a cycle loss function on CycleGAN.

Based on the knowledge of control systems \cite{nise2020control}, designing proper optimisers (or controllers) and advanced learning systems can benefit the learning process and complete relevant tasks (e.g., classification and generation). In this paper, we design two advanced optimisers and analyze three learning systems relying on the control system knowledge. The contributions are as follows: 

\textbf{Optimisers are controllers.}  \textbf{(1)} PID and  SGDM (PI controller) optimiser performs more stable than SGD (P controller), SGDM (PI controller), AdaM and fuzzyPID optimisers on most residual connection used CNN models. \textbf{(2)} HPF-SGD outperforms SGD and LPF-SGD, which indicates that high frequency part is significant during SGD learning process. \textbf{(3)} AdaM is an adaptive filter that combines an adaptive filter and an accumulation adaptive part.

\textbf{Learning systems of most ANNs are control systems.} \textbf{(1)} Most ANNs present perfect consistent performance with their system response. \textbf{(2)} We can use proper optimisers to control and improve the learning process of most ANNs. 

\textbf{The Optimiser should match the learning system.}  \textbf{(1)} RSs based vision models prefer SGDM, PID and fuzzyPID optimisers. \textbf{(2)} RS mechanism is similar to AdaM. particularly, SGDM optimizes the weight of models on the time dimension, and RS optimizes the model on the space dimension. \textbf{(3)} AdaM significantly benefits FFNN and GAN, but PID and FuzzyPID dotes CycleGAN most.

% Low-Pass-Filter SGD (LPFSGD) \cite{bisla2022low}

% Designing advanced learning systems can benefit the learning process and complete relevant tasks \cite{hinton2022forward}. tificial neural networks (ANNs) need to map the input $x$ to the output $y$ through parameters $\theta$. To measure the gap between the ANN output and the desired output, the backward training with the loss function $L$ is introduced. Given some training data, we can calculate the loss function $L(\theta, X_{train})$. In order to minimize the loss function $L$, we find the derivative of the loss function $L$ with respect to the parameter $\theta$ and update $\theta$ with the gradient descent method in most cases. However, the backpropagation learning strategy cannot be like the perceptual system which performs inference and learning in real-time without stopping \cite{hinton2022forward}. 

% Atificial neural networks (ANNs) need to map the input $x$ to the output $y$ through parameters $\theta$. To measure the gap between the ANN output and the desired output, the backward training with the loss function $L$ is introduced. Given some training data, we can calculate the loss function $L(\theta, X_{train})$. In order to minimize the loss function $L$, we find the derivative of the loss function $L$ with respect to the parameter $\theta$ and update $\theta$ with the gradient descent method in most cases. However, the backpropagation learning strategy cannot be like the perceptual system which performs inference and learning in real-time without stopping \cite{hinton2022forward}. 

\section{Problem Statement and Preliminaries}

\begin{wrapfigure}{r}{0.5\textwidth}
\centering
\includegraphics[scale=0.5]{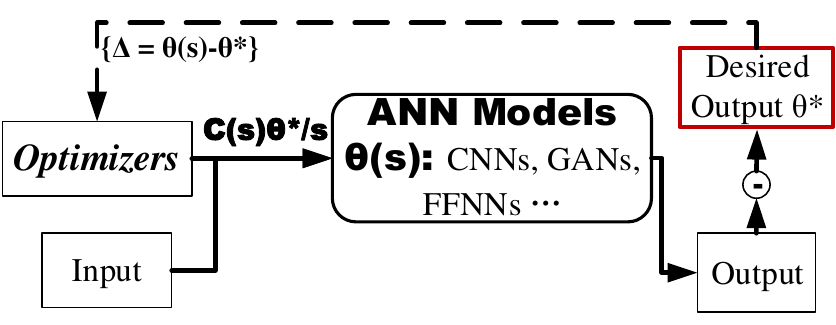}
    \captionof{figure}{The schematic structure of training ANN models. C(s) is the controller to train the target ANN model.}
\label{Figure1}
\end{wrapfigure}

% Even though we do not know whether these ANNs are linear or non-linear systems, fuzzy PID controllers always perform better than the classical PID controller in terms of convergence speed and precision. 

To make ANNs more effective and adaptive to specific tasks, controlling ANNs has become necessary. We initialize a parameter of a node in the ANN model as a scalar $\theta_{0}$. After enough time of updates, the optimal value of $\theta^{\ast}$ can be obtained. We simplify the parameter update in ANN optimisation as a one-step response (from $\theta_{0}$ to $\theta^{\ast}$ ) in the control system. The Laplace transform of $\theta^{\ast}$ is $\theta^{\ast}/s$. We denote the weight $\theta(t)$ at iteration $t$. The Laplace transform of $\theta(t)$ is denoted as $\theta(s)$, and that of error $e(t)=\theta^{\ast}-\theta(t)$ as $E(s)$:

\begin{equation} \label{Eq1}
E(s)=\frac{\theta^\ast}{s}-\theta(s)
\end{equation}

\noindent Considering the collaboration of backward and forward algorithms, the Laplace transform of the training process is
\begin{equation} \label{Eq2}
U(s)= \left(\mathit{Controller1}+\mathit{Controller2} \cdot F(s)\right) \cdot E(s)
\end{equation}

\noindent $F(s)$ is the forward system which has the capability to affect $U(s)$ beforehand. In our case, $u(t)$ corresponds to the update of $\theta(t)$. $\mathit{Controller1}$ is the parameter update algorithm for the backward process, and $\mathit{Controller2}$ is the parameter update algorithm for the forward process. Therefore, we replace $U(s)$ with $\theta(s)$ and $E(s)$ with $\left(\theta^\ast / s\right)-\theta(s)$. Equation \ref{Eq2} can be rewritten as
\begin{equation} \label{Eq3}
\theta(s) = \left(\mathit{Controller1}+\mathit{Controller2} \cdot F(s)\right) \cdot \left(\frac{\theta^{\ast}}{s}-\theta(s)\right)
\end{equation}

\noindent Finally, we simplify the formula of training a model as: 
\begin{equation} \label{Eq4}
\theta(s) = \frac{\mathit{Controller}}{\mathit{Controller}+1} \cdot \frac{\theta^{\ast}}{s}
\end{equation}

\noindent where $\mathit{Controller}=\mathit{Controller1}+\mathit{Controller2} \cdot F(s)$. $\theta^{\ast}$ denotes the optimal model which we should get at the end. Simplifying $\theta(s)$ further as below:
\begin{equation} \label{Eq5}
\theta(s) = \mathbf{Controller(s)} \cdot \mathbf{C(s)}
\end{equation}

\noindent where $\mathbf{Controller(s)}=\mathit{Controller}/(\mathit{Controller}+1)$, and $\mathbf{C(s)}=\theta^{\ast}/s$. Based on above analytic thought, as shown in Figure \ref{Figure1} there are two ways to obtain an optimal $\theta(s)$ and to make the training process better: \textbf{(1)} using a better \textbf{Controller} and \textbf{(2)} constructing a better training or control system $\mathbf{C(s)}$.

\section{Optimisers are Controllers}

In this section, we review several widely used optimisers, such as SGD \cite{robbins1951stochastic,cotter2011better,zhou2017convergence}, SGDM \cite{qian1999momentum,liu2020improved}, AdaM \cite{kingma2014adam,bock2018improvement}, PID-optimiser \cite{wang2020pid} and Gaussian LPF-SGD \cite{bisla2022low}. In the training process of most ANNs, there are diverse architectures used to satisfy various tasks. We analyze the performance of optimisers in terms of one node of backpropagation based ANN models. Please see the 
proof in Appendix A.

\subsection{AdaM Optimiser}

AdaM \cite{kingma2014adam} has been used to optimise the learning process of most ANNs, such as GAN, VAE, Transformer-based models, and their variants. We simplify the learning system of using AdaM on ANNs as below:

\begin{equation} \label{Eq6}
\theta(s) = \frac{K_{p}s + K_{i}} {Ms^2 + (K_{p}-Mln\beta_{1})s+K_{i}} \cdot \frac{\theta^\ast}{s}\\
\end{equation}

where $M$ is an adaption factor which will dynamically adjust the learning during the training process, and it can be derived from:

\begin{equation} \label{Eq7}
M = \frac{1}{\sqrt{\frac{ \sum_{i=0}^{t}\beta _{2}^{t-i}( \partial L_t / \partial \theta_t)^{2} }{ \sum_{i=0}^{t}\beta_{2}^{i-1} }} + \epsilon } \cdot \frac{ 1 }{ \sum_{i=0}^{t} \beta_{1}^{i-1} }
\end{equation}

Apart from the adaption part $M$, AdaM can be thought as the combination of SGDM and an adaptive filter with the cutoff frequency $\omega_c=ln(\beta_{1})$.

\subsection{Filter Processed SGD optimiser}

SGD learning process can be filtered under carefully designed filters. GLPF-SGD \cite{bisla2022low} used a low pass Gaussian-filter to smooth the training process, as well as actively searching the flat regions in the Deep Learning (DL) optimisation landscape. 
Eventually,  we simplify the learning system of using SGD with filters on ANNs as below:

\begin{equation} \label{Eq8}
\theta(s) = \frac{Gain \cdot \prod_{i=0}^{m} \left(s+h_{i}\right)}{Gain \cdot \prod_{i=0}^{m} \left(s+h_{i}\right) + \prod_{j=0}^{n} \left(s+l_{j}\right)} \cdot \frac{\theta^\ast}{s}
\end{equation}

where designed $Filter$ have the order, such as $n$ for the low pass and $m$ for the high pass ($h_{i}$ is the coefficient of the high pass part and $l_{i}$ is the coefficient of the low pass part), and $Gain$ is the gain factor:

\begin{equation} \label{Eq9}
Filter = Gain \cdot \frac{(s+h_{0})(s+h_{1})...(s+h_{m})}{(s+l_{0})(s+l_{1})...(s+l_{n})} 
\end{equation}

\subsection{PID and FuzzyPID optimiser}

Based on PID optimiser \cite{wang2020pid}, we design a PID controller which is optimised by fuzzy logic to make the training process more stable while keeping the dominant attribute of models. For instance, the ability to resist the disturbance of the poisoned samples, the quick convergent speed and the competitive performance. 

There are two key factors which affect the performance of the Fuzzy PID optimiser: (1) the selection of Fuzzy Universe Range $[-\varphi, \varphi]$ and (2) Membership Function Type $f_{m}$. 
\begin{gather} \label{eq10}
\begin{split}
& \widehat{K}_{\mathrm{P,I,D}}={K}_{\mathrm{P,I,D}}+\Delta K_{\mathrm{P,I,D}}
\end{split}
\end{gather}
\begin{gather} \label{eq11}
\begin{split}
& \Delta K_{\mathrm{P,I,D}} = Defuzzy(E(s),Ec(s)) \cdot K_{\mathrm{P,I,D}}\\
& Defuzzy(s) = f_{m}(round(-\varphi, \varphi, s))
\end{split}
\end{gather}

\noindent where $\Delta{K}_{\mathrm{P,I,D}}$ refer to the default gain coefficients of ${K}_{\mathrm{P}}$, ${K}_{\mathrm{I}}$ and ${K}_{\mathrm{D}}$ before modification. $E(s)$ is the back error, and $Ec(s)$ is the difference product between the $Laplace$ of $e(t)$ and $e(t-1)$. The Laplace function of this model $\theta(s)$ eventually becomes:
\begin{equation} \label{Eq12}
\theta(s) = \frac{\widehat{K}_{d}s^2+\widehat{K}_{p}s+\widehat{K}_{i}}{\widehat{K}_{d}s^2+(\widehat{K}_{p}+1)s+\widehat{K}_{i}} \cdot \frac{\theta^\ast}{s} 
\end{equation}
\noindent where $\widehat{K}_{p}$, $\widehat{K}_{i}$ and $\widehat{K}_{d}$ should be processed under the fuzzy logic. By carefully selecting the learning rate $r$, $\theta(s)$ becomes a stable system.

The PID \cite{ang2005pid} and Fuzzy PID \cite{tang2001optimal} controllers have been used to control a feedback system by exploiting the present, past, and future information of prediction error. The advantages of a fuzzy PID controller includes that it can provide different response levels to non-linear variations in a system. At the same time, the fuzzy PID controller can function as well as a standard PID controller in a system where variation is predictable.

\section{Control Systems of ANNs}

In this section, to systematically analyze the learning process of ANNs, we introduce three main common-used control systems that we believe can be respectively connected to backpropagation based CNNs, forward-forward algorithm based FFNNs, and GANs: \textbf{(1)} backward control system, \textbf{(2)} forward control system using different hyperparameters, and \textbf{(3)} backward-forward control system on different optimisers and hyperparameters. Please see the proof in Appendix B.

\subsection{Backward Control System}

Traditional CNNs use the backpropagation algorithm to update initialized weights, and based on errors or minibatched errors between real labels and predicted results, optimisers are used to control on how the weight should be updated. According to the deduction of PID optimiser \cite{wang2020pid}, the training process of Deep Neural Networks (DNNs) can be conducted under a step response of control systems. However, most common-used optimisers have their limitations, such as \textbf{(1)} SGD costs a very long term to reach convergence, \textbf{(2)} SGDM also has the side effect of long term convergence even with the momentum accelerating the training, \textbf{(3)} AdaM presents a frequent vibration during the training because of the merging of momentum and root mean squared propagation (RMSprop), \textbf{(4)} PID optimiser has better stability and convergence speed, but the training process is still vibrating. This proposed fuzzyPID optimiser can keep the learning process more stable, because it can be weighted towards types of responses, which seems like an adaptive gain setting on a standard PID optimiser. Finally, we get the system function $\theta(s)$ of ANNs by using FuzzyPID optimisers as an example below:
\begin{equation} \label{Eq13}
\theta(s) = \frac{\mathit{FuzzyPID}}{\mathit{FuzzyPID}+1} \cdot \frac{\theta^\ast}{s} 
\end{equation}

\subsection{Forward-Forward Control System}

The using of forward-forward computing algorithm was systematically analyzed in forward-forward neural network \cite{hinton2022forward} which aims to track features and figure out how ANNs can extract them from the training data. The Forward-Forward algorithm is a greedy multilayer learning procedure inspired by Boltzmann machines \cite{hinton1986learning} and noisy contrastive estimation \cite{gutmann2010noise}. To replace the forward-backward passes of backpropagation with two forward passes that operate on each other in exactly the same way, but on different data with opposite goals. In this system, the positive pass operates on the real data and adjusts the weights to increase the goodness in each hidden layer; the negative pass operates on the negative data and adjusts the weights to reduce the goodness in each hidden layer. According to the training process of FFNN, we get its system function $\theta(s)$ as below:
\begin{equation} \label{Eq14}
\theta(s) =  \left\{ \left( -(1 -\lambda)\frac{\theta^{\ast}}{s} + \lambda\frac{\theta^{\ast}}{s}-\left[ \theta(s)-\frac{Th}{s}\right] \right) \right\} \cdot Controller
\end{equation}

\noindent where $\lambda \in [0, 1]$ is the portion of positive samples, and $Th$ is the given Threshold according to the design \cite{hinton2022forward}. Input should contain negative and positive samples, and by adjusting the Threshold $Th$, the embedding space can be optimised. In each layer, weights should be updated on only corresponding errors that can be computed by subtracting the Threshold $Th$. We finally simplify $\theta(s)$ as:
\begin{equation} \label{Eq15}
\theta(s) = \frac{1}{Controller+1} \cdot \left( \frac{(2\lambda-1)\theta^{\ast}+Th}{s} \right) 
\end{equation}
\noindent Because $(2\lambda-1)\theta^{\ast} + Th \geq 0$, the system of FFNN is stable. Additionally, when $\lambda=0.5$ and $Th=1.0$, the learning system of FFNN (the second half part of Equation \ref{Eq15}) will become to that of backpropagation based CNN, as we assume $\theta^{\ast} \approx 1.0$. When $\lambda=0.5$, the optimal result $\theta^{\ast}$ has no relationship with the learning system.

\subsection{Backward-Forward Control System}

GAN is designed to generate samples from the Gaussian noise. The performance of the GAN depends on its architecture \cite{zhou2023gan}. The generative network uses random inputs to generate samples, and the discriminative network aims to classify whether the generated sample can be classified \cite{2014Generative}. We get its $\theta(s)$ as below:
\begin{equation} \label{Eq16}
\theta_D(s) = controller \cdot \theta_{G}(s) \cdot E(s)
\end{equation}
\begin{equation} \label{Eq17}
\theta_G(s) = controller \cdot E(s)
\end{equation} 
\begin{equation} \label{Eq18}
E(s) = \frac{\theta_{D}^{\ast}}{s} - \theta_{D}(s)
\end{equation} 
\noindent where $\theta_D(s)$ is the desired Discriminator, $\theta_G(s)$ is the desired Generator. $E(s)$ is the feed-back error. $\theta_{G}^{\ast}$ is the optimal solution of the generator, and $\theta_{D}^{\ast}$ is the optimal solution of the discriminator. 

Eventually, we simplify $\theta_G(s)$ and $\theta_D(s)$ as below:
\begin{equation} \label{Eq19}
\theta_G(s) = \frac{1}{2} \cdot \left( \frac{\theta_{D}^{\ast}}{Controller} \pm \sqrt{(\frac{\theta_D^\ast}{Controller})^{2} - \frac{4}{s}} \right)
\end{equation}
\begin{equation} \label{Eq20}
\theta_D(s) = \theta^{2}_G(s) 
\end{equation}
\noindent where if set $\theta_G(s)=0$, we get one pole point $s=0$. When using SGD as the $controller$, $\theta_G(s)$ is a marginally stable system.

\section{Experiments}

\subsection{Simulation} 

As we believe that the training process of most ANNs can be modeled as the source response of control systems, we use Simulink (MATLAB R2022a) to simulate their response to different sources. For the classification task, because all models aim to classify different categories,  we set a step source as illustrated in \cite{wang2020pid}. For the sample generation task, to get a clear generating result, we use a sinusoidal source.

% Two datasets (MNIST, CIFAR10) are used to evaluate the improvement of our advanced control systems that are applied to optimise the existing ANNs. 

\subsection{Experiment Settings} 

We train our models on the MNIST \cite{lecun1998gradient}, CIFAR10 \cite{krizhevsky2009learning}, CIFAR100 \cite{krizhevsky2009learning} and TinyImageNet \cite{le2015tiny} datasets. For an apple-to-apple comparison, our training strategy is mostly adopted from PID optimiser \cite{wang2020pid} and FFNN \cite{hinton2022forward}. To optimise the learning process, we \textbf{(1)} firstly use seven optimisers for the classification task on backpropagation algorithm based ANNs. \textbf{(2)} Secondly, we choose some important hyperparameters and simulate the learning process of FFNN. \textbf{(3)} Lastly, to improve the stability and convergence during the training of GAN, we analyze its system response on various optimisers. All models are trained on single Tesla V100 GPU. All the hyper-parameters are presented in Table \ref{Table3} of Appendix E.

% Our code is implemented based on DieT \cite{touvron2021training} \footnote{https://github.com/facebookresearch/deit} and CyclePLM \cite{chen2022cyclemlp} \footnote{https://github.com/ShoufaChen/CycleMLP}.

\subsubsection{Backward Control System} 
We design one neural network using backpropagation algorithm with $2$ hidden layers, setting the learning rate $r$ at $0.02$ and the fuzzy universe range $\varphi$ at $[-0.02, 0.02]$. We initialize $K_{P}$ as $1$, $K_{I}$ as $5$, and $K_{D}$ as $100$. Thus, we compare seven different optimisers: SGD (P controller), SGDM (PI controller), AdaM (PI controller with an Adaptive Filter), PID (PID controller), LPF-SGD, HPF-SGD and FuzzyPID (fuzzy PID controller) on the above ANN model. We set Gaussian membership function as the default membership function. See filter coefficients in Table \ref{Table4} of Appendix E. In Table \ref{Table5} of Appendix E, there is a set of hyperparameters  that we have used to trian CIFAR10, CIFAR100 and TinyImageNet.

\subsubsection{Forward-Forward Control System}  
Following the forward-forward algorithm \cite{hinton2022forward}, we design one forward-forward neural network (FFNN) with $4$ hidden layers each containing $2000$ ReLUs and full connectivity between layers, by simultaneously feeding positive and negative samples into the model to teach it to distinguish the handwriting number (MNIST). We also carefully select the proportion of positive and negative samples. The length of every block is $60$.

\subsubsection{Backward-Forward Control System} 
To demonstrate the relationship between the control system and the learning process of some complex ANNs, we choose the classical GAN \cite{2014Generative}. Both the generator and the discriminator comprise $4$ hidden layers. To verify the influence of different optimisers on GAN, we employ SGD, SGDM, AdaM, PID, LPF-SGD, HPF-SGD and fuzzyPID to generate the handwriting number (MNIST). We set the learning rate at $0.0002$ and the total number of epochs at $200$.

\section{Results and Analysis}

In this section, we present simulation performance, classification accuracy, error rate and generation result, using different optimisers and advanced control systems.

\subsection{Backward Control System on CNN}

\begin{figure}[h]
    \centering 
    \begin{subfigure}[b]{0.32\textwidth} 
        \centering 
        \includegraphics[scale=0.3]{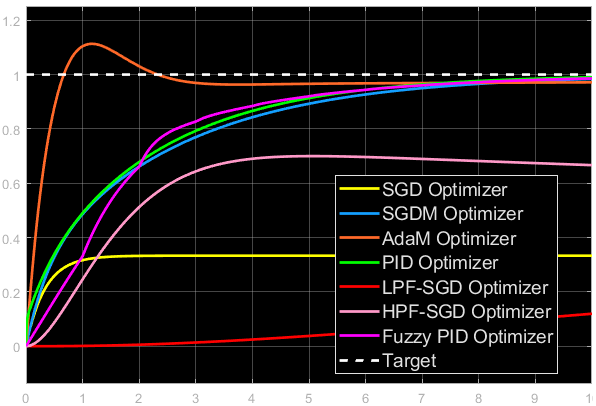} \caption{\small The step response of CNN using different controllers (or optimisers). }
        \label{Figure2-a} 
    \end{subfigure} 
    \hfill 
    \begin{subfigure}[b]{0.32\textwidth} 
        \centering 
        \includegraphics[scale=0.3]{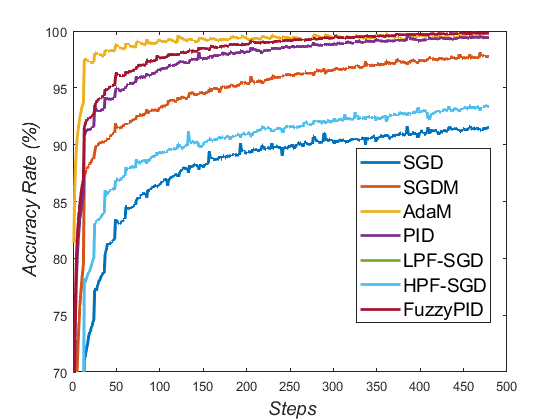} \caption{\small The training curve of CNN using different optimisers (or controllers) on MNIST. }
        \label{Figure2-b} 
    \end{subfigure} 
    \hfill 
    \begin{subfigure}[b]{0.32\textwidth}
        \centering 
        \includegraphics[scale=0.3]{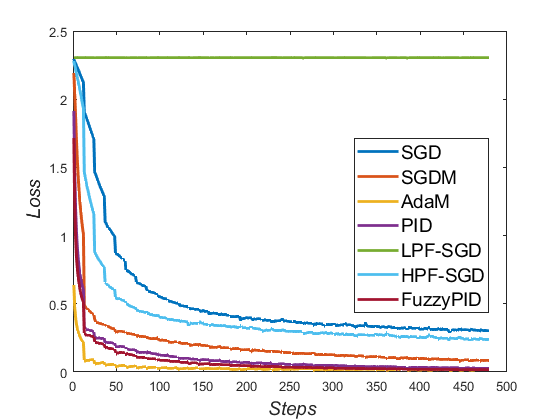} 
        \caption{\small  The loss curve of CNN using different optimisers (or controllers) on MNIST.} 
        \label{Figure2-c} 
    \end{subfigure}
    \caption{The step response, training curve and loss curve using different controllers, such as SGD, SGDM, AdaM, PID, LPF-SGD, HPF-SGD and FuzzyPID optimisers.} 
    \label{Figure2} 
\end{figure}

\begin{table}[h]
\caption{The results of ANN based on the backpropogation algorithm on MNIST data. Using the 10-fold cross-validation, the average and standard variance results are shown below. }
\centering
\label{Table1}
\scriptsize
\begin{tabular}{p{0.16\linewidth} p{0.07\linewidth} p{0.07\linewidth} p{0.07\linewidth} p{0.07\linewidth} p{0.08\linewidth} p{0.08\linewidth} p{0.09\linewidth}}
\toprule
\textbf{optimiser}  &SGD  &SGDM  &Adam  &PID   &LPF-SGD  &HPF-SGD 
  &FuzzyPID  \\
\hline
\textbf{Training $Accuracy$} & $91.48_{\pm0.03}$  &$97.78_{\pm0.00}$   &$99.46_{\pm0.02}$  &$99.45_{\pm0.01}$   &$11.03_{\pm0.01}$   &$93.35_{\pm0.02}$  &$99.73_{\pm0.09}$  \\
\textbf{Testing $Accuracy$}  &$91.98_{\pm0.05}$   &$97.11_{\pm0.02}$  
 &$97.81_{\pm0.10}$  &$98.18_{\pm0.02}$  &$10.51_{\pm0.03}$  &$93.45_{\pm0.09}$  & $98.24_{\pm0.10}$  \\ 
\bottomrule
\end{tabular}
\end{table}

Before doing the classification task, we firstly simulate the step response of backpropagation based ANNs on each controller (optimiser). As observed in Figure \ref{Figure2-b} and Figure \ref{Figure2-c}, AdaM optimiser can rapidly converge to the optimal but with an obvious vibration. Although FuzzyPID cannot rapidly converge to the optimal, there is no obvious vibration during the training. Other optimisers, such as HPF-SGD, SGDM and PID, perform lower than AdaM and FuzzyPID in terms of the training process. In Figure \ref{Figure2-a}, the response of AdaM controller is faster than others, and FuzzyPID follows it. However, due to the overshoot on AdaM, the stability of ANN system when using the AdaM controller tends to be lower. This overshoot phenomenon is reflected on the training process of Adam optimising in Figure \ref{Figure2-b} and Figure \ref{Figure2-c}.

We summarize the result of classifying MNIST in Table \ref{Table1}. Under the same condition, SGD optimiser reaches the testing accuracy at $91.98\%$, but other optimisers can reach above $97\%$. FuzzyPID gets the highest training and testing accuracy rates using Guassian membership function. In Figure \ref{Figure2}, if considering the rise time, the settling time and the overshoot, the fuzzy optimiser outperforms other optimisers. A better optimiser (or controller) that has inherited advanced knowledge and sometimes has been effectively designed is beneficial for the classification performance.

\subsection{Forward Forward Control System on FFNN}

\begin{figure}[h]
    \centering 
    \begin{subfigure}[b]{0.24\textwidth}
        \centering 
        \includegraphics[scale=0.22]{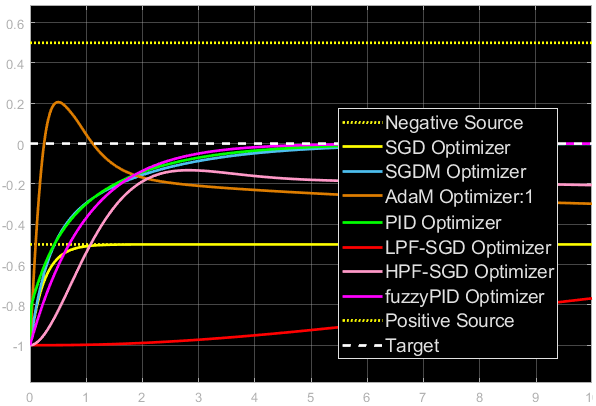} 
        \caption{\small The system response of FFNN on corresponding optimisers.} 
        \label{Figure3-a} 
    \end{subfigure} 
    \begin{subfigure}[b]{0.24\textwidth}
        \centering 
        \includegraphics[scale=0.23]{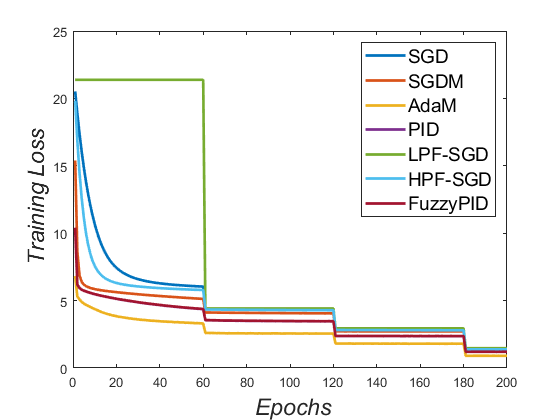} 
        \caption{\small The loss curve of FFNN on corresponding optimisers.} 
        \label{Figure3-b} 
    \end{subfigure}
    \begin{subfigure}[b]{0.24\textwidth}
        \centering 
        \includegraphics[scale=0.22]{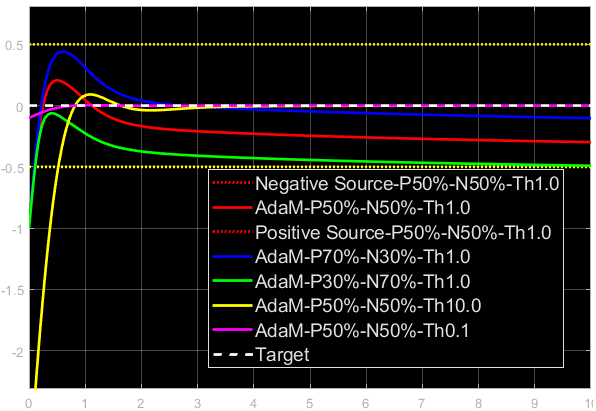} 
        \caption{\small The system response of FFNN on corresponding hyperparameters.} 
        \label{Figure3-c} 
    \end{subfigure}
    \begin{subfigure}[b]{0.24\textwidth}
        \centering 
        \includegraphics[scale=0.23]{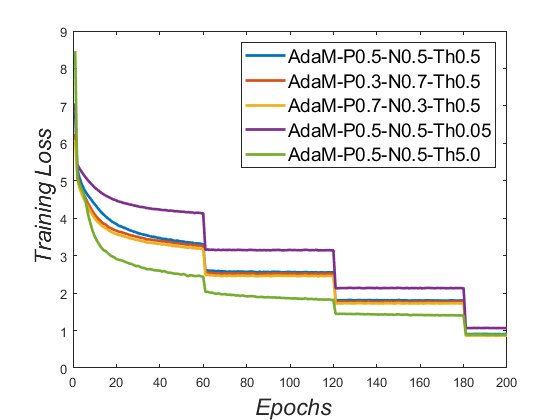} 
        \caption{\small The loss curve of FFNN on corresponding hyperparameters.} 
        \label{Figure3-d} 
    \end{subfigure}
    \caption{The step response and loss curve of FFNN using different controllers and various hyperparameters.} 
    \label{Figure3} 
\end{figure}

\begin{table}[h]
\caption{The error rate (\%) of FFNN using different optimisers and various hyperparameters on MNIST. Using the 10-fold cross-validation, the average and standard variance results are shown below.}
\label{Table2}
\centering
\scriptsize
\begin{tabular}{p{0.08\linewidth} p{0.04\linewidth} p{0.04\linewidth} p{0.04\linewidth} p{0.04\linewidth} 
p{0.05\linewidth} p{0.05\linewidth} p{0.05\linewidth} p{0.05\linewidth} p{0.05\linewidth} p{0.05\linewidth} p{0.06\linewidth}}
\toprule
\textbf{Method}  & \begin{tabular}[c]{@{}l@{}}50\% P,\\ 50\% N, \\ Th=1.0, \\ SGD\end{tabular} & \begin{tabular}[c]{@{}l@{}}50\% P, \\ 50\% N, \\ Th=1.0, \\ SGDM\end{tabular} & \begin{tabular}[c]{@{}l@{}}50\% P, \\ 50\% N, \\ Th=1.0, \\ Adam\end{tabular} & \begin{tabular}[c]{@{}l@{}}50\% P, \\ 50\% N, \\ Th=1.0, \\ PID\end{tabular} & \begin{tabular}[c]{@{}l@{}}50\% P, \\ 50\% N, \\ Th=1.0, \\ LPF-SGD\end{tabular} & \begin{tabular}[c]{@{}l@{}}50\% P, \\ 50\% N, \\ Th=1.0, \\ HPF-SGD\end{tabular}  & \begin{tabular}[c]{@{}l@{}}50\% P, \\ 50\% N, \\ Th=1.0, \\ FuzzyPID\end{tabular}& \begin{tabular}[c]{@{}l@{}}30\% P, \\ 70\% N, \\ Th=1.0, \\ Adam\end{tabular} & \begin{tabular}[c]{@{}l@{}}70\% P, \\ 30\% N, \\ Th=1.0, \\ Adam\end{tabular} & \begin{tabular}[c]{@{}l@{}}50\% P, \\ 50\% N, \\ Th=0.1, \\ Adam\end{tabular} & \begin{tabular}[c]{@{}l@{}}50\% P, \\ 50\% N, \\ Th=10.0, \\ Adam\end{tabular} \\ \hline
\begin{tabular}[c]{@{}l@{}}\textbf{Train Error}\end{tabular}                  & \begin{tabular}[c]{@{}l@{}}$68.96$\\$\tiny{\pm0.79}$\end{tabular}                                                                                                                                                                                      & \begin{tabular}[c]{@{}l@{}}$24.66$\\$\tiny{\pm 0.23}$\end{tabular}                                                                                             & \begin{tabular}[c]{@{}l@{}}$4.57$\\$\tiny{\pm0.23}$\end{tabular}                                                                                           & \begin{tabular}[c]{@{}l@{}}$14.90$\\$\tiny{\pm0.11}$\end{tabular}                                                                                                                                                                                       & \begin{tabular}[c]{@{}l@{}}$93.00$\\$\tiny{\pm0.15}$\end{tabular}                                                                                     & \begin{tabular}[c]{@{}l@{}}$48.24$\\$\tiny{\pm0.14}$\end{tabular}                                                                                           
& \begin{tabular}[c]{@{}l@{}}$14.96$\\$\tiny{\pm0.19}$\end{tabular}                                                                                            & \begin{tabular}[c]{@{}l@{}}$4.82$\\$\tiny{\pm0.09}$\end{tabular}                                                                                                                                                                                        & \begin{tabular}[c]{@{}l@{}}$3.61$\\$\tiny{\pm0.08}$\end{tabular}                                                                                                                                                                                                                                                            & \begin{tabular}[c]{@{}l@{}}$6.44$\\$\tiny{\pm0.10}$\end{tabular}                         & \begin{tabular}[c]{@{}l@{}}$1.15$\\$\tiny{\pm0.05}$\end{tabular}         \\ \hline
\begin{tabular}[c]{@{}l@{}}\textbf{Test Error}\end{tabular}                  & \begin{tabular}[c]{@{}l@{}}$68.85$\\$\tiny{\pm0.90}$\end{tabular}                                                                                          & \begin{tabular}[c]{@{}l@{}}$24.02$\\$\tiny{\pm0.30}$\end{tabular}                                                                                                                                                                                      & \begin{tabular}[c]{@{}l@{}}$5.00$\\$\tiny{\pm0.30}$\end{tabular}                                                                                                                                                                                                                                                                                & \begin{tabular}[c]{@{}l@{}}$14.38$\\$\tiny{\pm0.15}$\end{tabular}                                                                                                                                                                                                                                                                                                                                                                          & \begin{tabular}[c]{@{}l@{}}$92.89$\\$\tiny{\pm0.36}$\end{tabular}                                                                                                                                                                                                                                                                                                                                                                                                                                                              
& \begin{tabular}[c]{@{}l@{}}$48.45$\\$\tiny{\pm0.23}$\end{tabular}                                                                                                                                                                                                                                                                                                                                                                                                                                                                     & \begin{tabular}[c]{@{}l@{}}$14.43$\\$\tiny{\pm0.25}$\end{tabular}                                                                                                                                                                                                                                                                                                                                                                                                                                                                      & \begin{tabular}[c]{@{}l@{}}$5.31$\\$\tiny{\pm0.13}$\end{tabular}                                                                                                                                                                                                                                                                                                                                                                                                                                                                                                                                                                  & \begin{tabular}[c]{@{}l@{}}$4.37$\\$\tiny{\pm0.11}$\end{tabular}                                                                                                                                                                                                                                                                                                                                                                                                                                                                                                                                                                                                                                      & \begin{tabular}[c]{@{}l@{}}$6.52$\\$\tiny{\pm0.13}$\end{tabular}                                                                                                                                                                                                                                                                                                                                                                                                                                                                      & \begin{tabular}[c]{@{}l@{}}$1.35$\\$\tiny{\pm0.08}$\end{tabular}                                                                                                                                                                                                                                                                                                                                                                                                                                                                              \\ \bottomrule
\end{tabular}
\end{table}

We also simulate the control system of this proposed FFNN and compare its system response on different hyperparameters. In Figure \ref{Figure3}, SGD controller still cannot reach the target, and AdaM controller reacts fastest approaching to the target. However, SGDM controller lags behind PID in terms of the step response. Because of the low frequency part of LPF-SGD, it climbs slower than HPF-SGD. Although the differential coefficient D of PID optimiser can help reduce overshoot and overcome oscillation and reduce the adjustment time, its performance cannot catch up with AdaM. Compared to Table \ref{Table2}, AdaM outperforms other optimisers in terms of error rates, and the performance of these seven optimisers are echoing Figure \ref{Figure3-a}. A higher portion of positive samples can contribute to the classification, and a higher $Threshold$ can benefit more. For the step response in Figure \ref{Figure3-c}, although AdaM ( $Threshold=0.5$, $portion$ of positive samples is $70\%$, and $portion$ of negative samples is $30\%$) and AdaM ( $Threshold=0.5$, $portion$ of positive samples is $50\%$, and $portion$ of negative samples is $50\%$) rise fatest, the final results in Table \ref{Table2} present that AdaM ( $Threshold=5.0$, $portion$ of positive samples is $50\%$, and $portion$ of negative samples is $50\%$) get a lower error rate.

\subsection{Backward-Forward Control System on GAN}

    %\vskip 
    %\baselineskip 

\begin{wrapfigure}{r}{0.5\textwidth}
    \centering 
    \begin{subfigure}[b]{0.12\textwidth} 
        \caption{\small The first epoch.} 
        \centering 
        \includegraphics[scale=0.18]{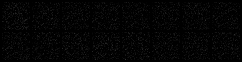} 
        % \label{fig:mean and std of net14} 
    \end{subfigure} 
    \hfill 
    \begin{subfigure}[b]{0.12\textwidth} 
        \caption{\small The $50_{th}$ epoch.} 
        \centering 
        \includegraphics[scale=0.18]{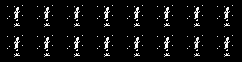}
        % \label{fig:mean and std of net24} 
    \end{subfigure}
    \hfill
    \begin{subfigure}[b]{0.12\textwidth} 
        \caption{\small The $100_{th}$ epoch.}
        \centering 
        \includegraphics[scale=0.18]{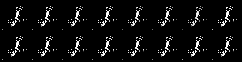}
        % \label{fig:mean and std of net14} 
    \end{subfigure} 
    \hfill 
    \begin{subfigure}[b]{0.12\textwidth} 
        \caption{\small The $200_{th}$ epoch.}
        \centering 
        \includegraphics[scale=0.18]{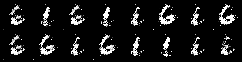}
        % \label{fig:mean and std of net24} 
    \end{subfigure} 

    \begin{subfigure}[b]{0.12\textwidth} 
        \centering 
        \includegraphics[scale=0.18]{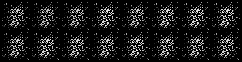} 
        % \caption{\small 1}
        % \label{fig:mean and std of net14} 
    \end{subfigure} 
    \hfill 
    \begin{subfigure}[b]{0.12\textwidth} 
        \centering 
        \includegraphics[scale=0.18]{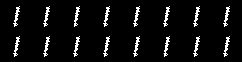} 
        % \caption{\small 50} 
        % \label{fig:mean and std of net24} 
    \end{subfigure}
    \hfill
    \begin{subfigure}[b]{0.12\textwidth} 
        \centering 
        \includegraphics[scale=0.18]{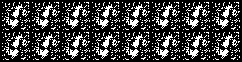} 
        % \caption{\small 100} 
        % \label{fig:mean and std of net14} 
    \end{subfigure} 
    \hfill 
    \begin{subfigure}[b]{0.12\textwidth} 
        \centering 
        \includegraphics[scale=0.18]{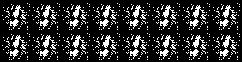} 
        % \caption{\small 150} 
        % \label{fig:mean and std of net24} 
    \end{subfigure} 

    \begin{subfigure}[b]{0.12\textwidth} 
        \centering 
        \includegraphics[scale=0.18]{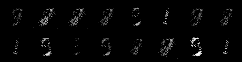} 
        % \caption{\small 1}
        % \label{fig:mean and std of net14} 
    \end{subfigure} 
    \hfill 
    \begin{subfigure}[b]{0.12\textwidth} 
        \centering 
        \includegraphics[scale=0.18]{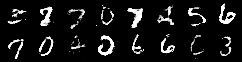} 
        % \caption{\small 50} 
        % \label{fig:mean and std of net24} 
    \end{subfigure}
    \hfill
    \begin{subfigure}[b]{0.12\textwidth} 
        \centering 
        \includegraphics[scale=0.18]{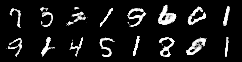} 
        % \caption{\small 100} 
        % \label{fig:mean and std of net14} 
    \end{subfigure} 
    \hfill 
    \begin{subfigure}[b]{0.12\textwidth} 
        \centering 
        \includegraphics[scale=0.18]{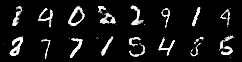} 
        % \caption{\small 150} 
        % \label{fig:mean and std of net24} 
    \end{subfigure} 

    \begin{subfigure}[b]{0.12\textwidth} 
        \centering 
        \includegraphics[scale=0.18]{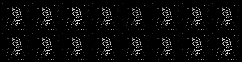} 
        % \caption{\small 1}
        % \label{fig:mean and std of net14} 
    \end{subfigure} 
    \hfill 
    \begin{subfigure}[b]{0.12\textwidth} 
        \centering 
        \includegraphics[scale=0.18]{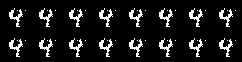} 
        % \caption{\small 50} 
        % \label{fig:mean and std of net24} 
    \end{subfigure}
    \hfill
    \begin{subfigure}[b]{0.12\textwidth} 
        \centering 
        \includegraphics[scale=0.18]{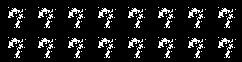} 
        % \caption{\small 100} 
        % \label{fig:mean and std of net14} 
    \end{subfigure} 
    \hfill 
    \begin{subfigure}[b]{0.12\textwidth} 
        \centering 
        \includegraphics[scale=0.18]{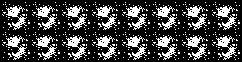} 
        % \caption{\small 150} 
        % \label{fig:mean and std of net24} 
    \end{subfigure}\

    \begin{subfigure}[b]{0.12\textwidth} 
        \centering 
        \includegraphics[scale=0.18]{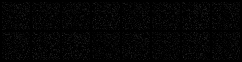} 
        % \caption{\small 1}
        % \label{fig:mean and std of net14} 
    \end{subfigure} 
    \hfill 
    \begin{subfigure}[b]{0.12\textwidth} 
        \centering 
        \includegraphics[scale=0.18]{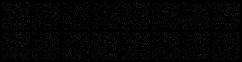} 
        % \caption{\small 50} 
        % \label{fig:mean and std of net24} 
    \end{subfigure}
    \hfill
    \begin{subfigure}[b]{0.12\textwidth} 
        \centering 
        \includegraphics[scale=0.18]{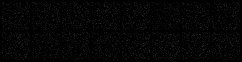} 
        % \caption{\small 100} 
        % \label{fig:mean and std of net14} 
    \end{subfigure} 
    \hfill 
    \begin{subfigure}[b]{0.12\textwidth} 
        \centering 
        \includegraphics[scale=0.18]{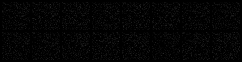} 
        % \caption{\small 150} 
        % \label{fig:mean and std of net24} 
    \end{subfigure} 

    \begin{subfigure}[b]{0.12\textwidth} 
        \centering 
        \includegraphics[scale=0.18]{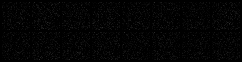} 
        % \caption{\small 1}
        % \label{fig:mean and std of net14} 
    \end{subfigure} 
    \hfill 
    \begin{subfigure}[b]{0.12\textwidth} 
        \centering 
        \includegraphics[scale=0.18]{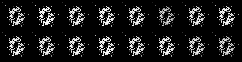} 
        % \caption{\small 50} 
        % \label{fig:mean and std of net24} 
    \end{subfigure}
    \hfill
    \begin{subfigure}[b]{0.12\textwidth} 
        \centering 
        \includegraphics[scale=0.18]{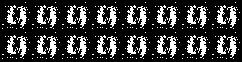} 
        % \caption{\small 100} 
        % \label{fig:mean and std of net14} 
    \end{subfigure} 
    \hfill 
    \begin{subfigure}[b]{0.12\textwidth} 
        \centering 
        \includegraphics[scale=0.18]{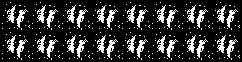} 
        % \caption{\small 150} 
        % \label{fig:mean and std of net24} 
    \end{subfigure} 

    \begin{subfigure}[b]{0.12\textwidth} 
        \centering 
        \includegraphics[scale=0.18]{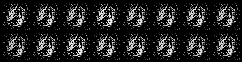} 
        % \caption{\small 1}
        % \label{fig:mean and std of net14} 
    \end{subfigure} 
    \hfill 
    \begin{subfigure}[b]{0.12\textwidth} 
        \centering 
        \includegraphics[scale=0.18]{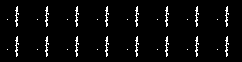} 
        % \caption{\small 50} 
        % \label{fig:mean and std of net24} 
    \end{subfigure}
    \hfill
    \begin{subfigure}[b]{0.12\textwidth} 
        \centering 
        \includegraphics[scale=0.18]{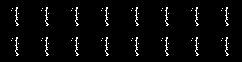} 
        % \caption{\small 100} 
        % \label{fig:mean and std of net14} 
    \end{subfigure} 
    \hfill 
    \begin{subfigure}[b]{0.12\textwidth} 
        \centering 
        \includegraphics[scale=0.18]{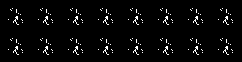} 
        % \caption{\small 150} 
        % \label{fig:mean and std of net24} 
    \end{subfigure} 
    \caption{The generated samples from classical GAN on corresponding optimisers (from top to bottom is respectively SGD, SGDM, AdaM, PID, LPF-SGD, HPF-SGD, and FuzzyPID).} 
    \label{Figure4} 
\end{wrapfigure}

\begin{figure}[h]
    \centering 
    \begin{subfigure}[b]{0.13\textwidth} 
        \centering 
        \includegraphics[scale=0.13]{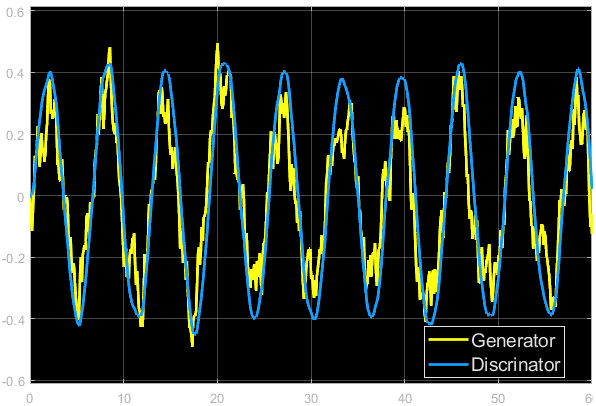} 
        %\caption{\small Classical GAN on SGD optimiser.} 
        \label{Figure5-a} 
    \end{subfigure} 
    \begin{subfigure}[b]{0.13\textwidth}
        \centering 
        \includegraphics[scale=0.13]{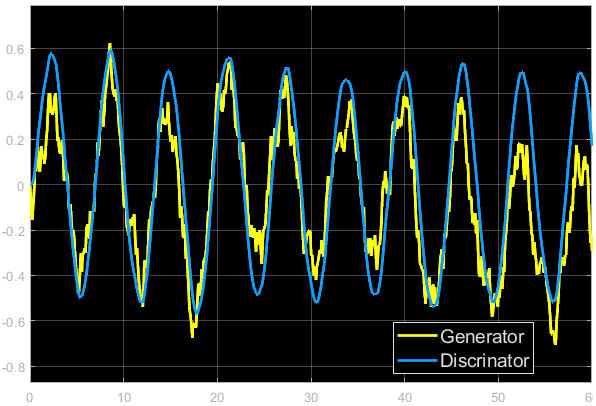} 
        %\caption{\small Classical GAN on SGDM optimiser.} 
        \label{Figure5-b} 
    \end{subfigure}
    \begin{subfigure}[b]{0.13\textwidth}
        \centering 
        \includegraphics[scale=0.13]{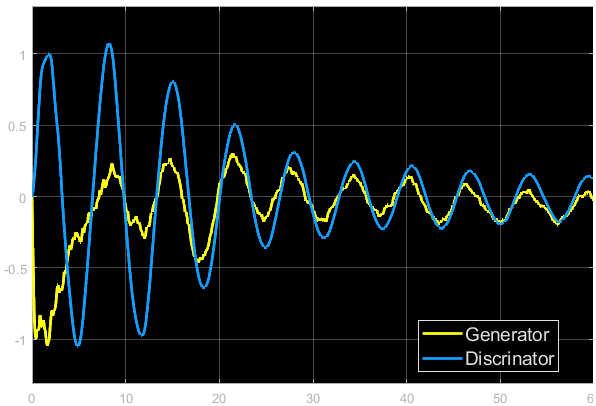} 
        %\caption{\small  Classical GAN on AdaM optimiser.} 
        \label{Figure5-c} 
    \end{subfigure}
    \begin{subfigure}[b]{0.13\textwidth}
        \centering 
        \includegraphics[scale=0.13]{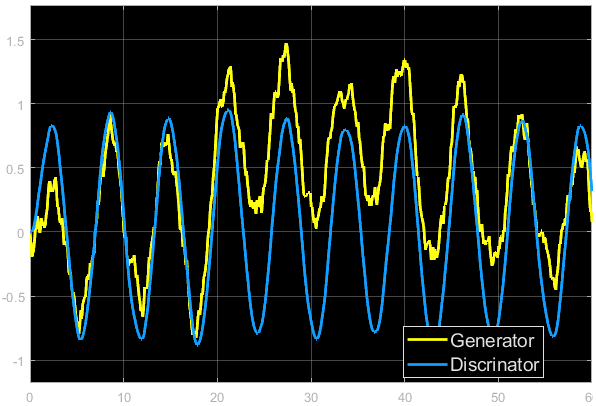} 
        %\caption{\small  Classical GAN on PID optimiser.} 
        \label{Figure5-d} 
    \end{subfigure}
    \begin{subfigure}[b]{0.13\textwidth} 
        \centering 
        \includegraphics[scale=0.13]{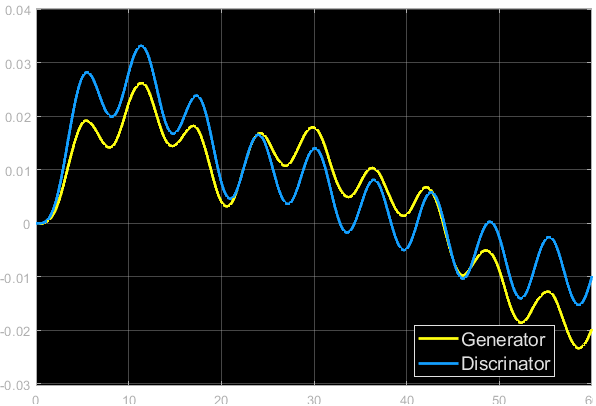} 
        %\caption{\small Classical GAN on LPF-SGD optimiser.} 
        \label{Figure5-e} 
    \end{subfigure} 
    \begin{subfigure}[b]{0.13\textwidth} 
        \centering 
        \includegraphics[scale=0.13]{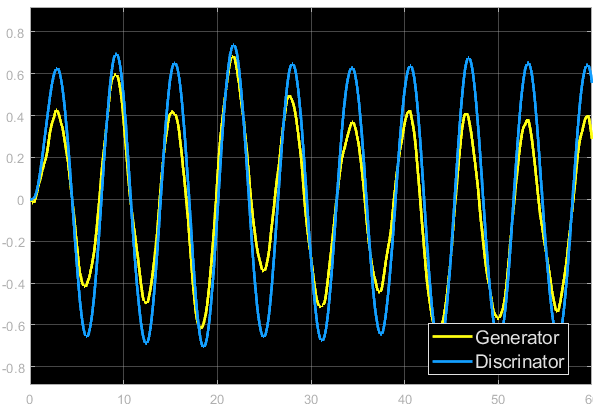} 
        %\caption{\small Classical GAN on HPF-SGD optimiser.} 
        \label{Figure5-f} 
    \end{subfigure} 
    \begin{subfigure}[b]{0.13\textwidth} 
        \centering 
        \includegraphics[scale=0.13]{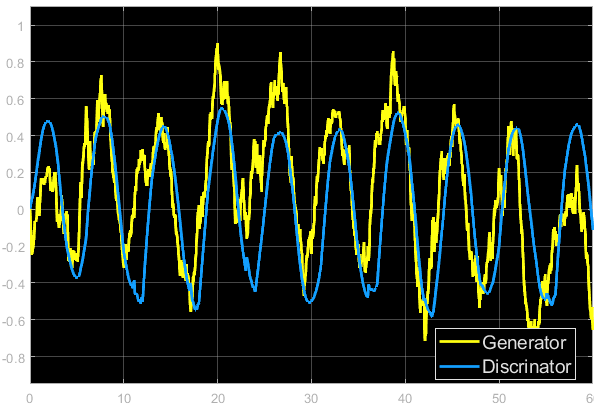} 
        %\caption{\small Classical GAN on FuzzyPID optimiser.} 
        \label{Figure5-g} 
    \end{subfigure} 
    \caption{The system response of Classical GAN on different hyperparameters and various optimisers. Optimiser from left to right is respectively SGD, SGDM, AdaM, PID, LPFSGD, HPFSGD, and FuzzyPID. (Blue is the discriminator, and yellow is the generator)} 
    \label{Figure5} 
\end{figure}

For the sample generation task, we also simulate the system response of GANs on each controllers (optimisers) and summarize the result in Figure \ref{Figure5}. Apart from AdaM, LPF-SGD and HPF-SGD, all controllers have obvious noise, and interestingly, this phenomenon can be seen in Figure \ref{Figure4}. The generated MNIST using Adam optimiser has no noise and can be easily recognized, and not surprised, the source response of AdaM in Figure \ref{Figure5} can finally converge. Figure \ref{Figure4} and Figure \ref{Figure5} mutually echo each other. Eventually, when using classical GAN to generate samples, AdaM should be the best optimiser to optimise the update of weights. The generated MNIST sample sometimes cannot be recognized, and GAN generates only same samples. One reason for this can be observed in Figure \ref{Figure5}, where the sinusoidal signals generated by these four controllers, such as PID, LPF-SGD, HPF-SGD and FuzzyPID move up and down, potentially leading to an unstable and same generation output.

\section{Discussion}

\subsection{Why various optimisers are controllers during the learning process?}

Under the same training condition (e.g., same architecture and hyperparameters), corresponding optimisers can tackle with specific tasks. Residual connection used vision models prefer SGDM, HPF-SGD and PID optimisers (Seen from Figure \ref{Figure14} of Appendix F). There is an obvious overshoot on the step response of AdaM controller (Seen from Figure \ref{Figure9}), and a similar vibration can be found in the testing curve of Figure \ref{Figure14} of Appendix F. The classification task always needs a rapid response to save learning resources, but if stability and robustness are the priorities, we should set others as the opimizer, such as PID or FuzzyPID optimiser, which under fuzzy logic adjustment, demonstrates a superior step response (can be seen from Figure \ref{Figure2-a}). Moreover, for the generation task, GAN satisfies AdaM optimiser. We found that the adaptive part of AdaM can rapidly adjust the learning process. However, other optimisers, such as SGD, SGDM and PID, generate samples with obvious noise and output the same samples make the generated sample cannot be recognized easily (can be seen from Figure \ref{Figure4} and Figure \ref{Figure5}). For particular needs (e.g., Image-to-Image Translation), CycelGAN, this advanced generation system was proposed to generate samples from one data pool and to improve its domain adaption on the target data pool. Coincidentally, we found that CycleGAN has a preference for the PID optimiser. Therefore, it is necessary to design a stable and task-satisfied optimiser on a specificly designed learning system. However, given that the system functions of most learning systems are extremely complex, simulating their system responses has become a viable way to analyze them. We conclude that to achieve best performance, every ANN should use the proper optimiser according to its learning system.

\subsection{How various learning systems can be analyzed?}

Numerous advanced components have enhanced ANNs. Conducting a quantitative analysis on each of them can pave the way for the development of new optimisers and learning systems. For the classification task using a backward control system, in one node of the learning system, and in terms of analyzing a single component, the rise time, peak time, overshoot (vibration), and settling time \cite{wang2020pid,nise2020control} can be the metrics to evaluate the performance of such component on learning systems. To visualize the learning process, FFNN was proposed by \cite{hinton2022forward} , and effectively, this forward-forward-based training system also can achieve competitive performance compared to backpropagation-based models. The $Threshold$ -- one hyperparameter -- can significantly benefit the convergence speed, as it has the effect of proportional adjustment (same as a stronger P in PID controller). The portion of positive samples can slightly affect the classification result, as because the proportional adjustment is too weak on FFNN learning system (Seen from Equation \ref{Eq15}). Additionally, the system response on various sources can also serve as a metric to evaluate the learning system. We conclude that there are two main branches to improve ANNs: \textbf{(1)} develop a proper optimiser; \textbf{(2)} design a better learning system. On the one hand, for example, the system response of GAN has high-frequency noise and cannot converge using SGD, SGDM and PID optimisers (seen from Figure \ref{Figure5}). One possible solution is adding an adaptive filter. Thus, AdaM outperforms other optimisers on generating samples (Seen from Figure \ref{Figure4}). The overshoot of AdaM and SGDM during the learning process of classification tasks can accelerate the convergence, but its side-effect of vibration brings us to PID and FuzzyPID. Therefore, developing a task-matched optimiser according to the system response determines the final performance of ANNs. On the other hand, to satisfy various task requirements, learning systems also should become stable and fast. For example, $\theta_{G}(s)$ has two system functions as derived from Eq \ref{Eq19}), to offset the side effect by considering the possible way using extra generator. That can explain why other advanced GANs using multi-generators (e.g., CycleGAN) can generate high-quality samples than the classical GAN.

\section{Limitations}

Although we systematically proved that \textbf{(1)} the optimiser acts as a controller and \textbf{(2)} the learning system functions as a control system, in this preliminary work, there are three obvious limitations: \textbf{a.} we cannot analyze larger models due to the complexity introduced by advanced techniques; \textbf{b.} the system response of some ANNs (e.g., FFNN) may not perfectly align with their real performance; \textbf{c.} we cannot always derive the solution of complex learning system.

\section{Conclusion}

In this study, we showed comprehensive empirical study investigating the connection between control systems and various learning systems of ANNs. We provided a systematic analysis method for several ANNs, such as CNN, FFNN, GAN, CycleGAN, and ResNet on several optimisers: SGD, SGDM, AdaM, PID, LPF-SGD, HPF-SGD and FuzzyPID. By analyzing the system response of ANNs, we explained the rationale behind choosing appropriate optimisers for different ANNs. Moreover, designing better learning systems under the use of proper optimiser can satisfy task requirements. In our future work, we will intend to delve into the the control system of other ANNs, such as Variational Autoencoders (VAEs), diffusion models, Transformer-based models and so on, aw well as the development of optimisers, as we believe the principles of control systems can guide improvements in all ANNs and optimisers.

%----------------------------------------------------------
%               HERE IS EXAMPLE OF A FIGURE	
%		environment {figure} for one-column figure, {figure\ast} for two-column figure 
%	 	\caption{} for one-column caption, \captionwide{} for two-column figure
%
% \begin{figure}[ht!]
% \begin{center}
% \resizebox{65mm}{!}{\includegraphics{obr1.eps}}
% %\input{obr/dr1a.pstex_t}
% \caption{The description of a figure is of the same style as the description of a table; the figure itself is of the environment \texttt{figure}.
% } 
% \label{fig1}
% \end{center}
% \end{figure}%\vspace{5mm} %Hannes Nordmann
%
%		example of twocolumn figure - pay attention to figure numbering, may be wrong.
%
%\begin{figure\ast}[ht!]
%\begin{center}
%\resizebox{65mm}{!}{\includegraphics{obr1.eps}}
%\captionwide{The example of two-column figure and caption.
%} 
%\label{figwide}
%\end{center}
%\end{figure\ast}
%

% %----------------------------------------------------------
% %               THIS IS THE PLACE FOR  ACKNOWLEDGEMENTS
% \section\ast{Acknowledgements}
% The research described in the paper was supervised by  doc. Ing. Daniel Novák, Ph.D, FEE CTU in Prague and supported by the Research Centre for Informatics (grant NO. $CZ.02.1.01/0.0/16\_019/0000765$).

%----------------------------------------------------------
%               HERE WRITE YOUR PAPER

%Bibliography
\bibliographystyle{unsrt}  
\bibliography{templateArxiv}

% \clearpage
%%%%%%%%%%%%%%%%%%%%%%%%%%%%%%%%%%%%%%%%%%%%%%%%%%%%%%%%%%%%


%%%%%%%%%%%%%%%%%%%%%%%%%%%%%%%%%%%%%%%%%%%%%%%%%%%%%%%%%%%%%%%%%%%%%%%%%%%%%%%
%%%%%%%%%%%%%%%%%%%%%%%%%%%%%%%%%%%%%%%%%%%%%%%%%%%%%%%%%%%%%%%%%%%%%%%%%%%%%%%

%%%%%%%%%%%%%%%%%%%%%%%%%%%%%%%%%%%%%%%%%%%%%%%%%%%%%%%%%%%%%%%%%%%%%%%%%%%%%%%
%%%%%%%%%%%%%%%%%%%%%%%%%%%%%%%%%%%%%%%%%%%%%%%%%%%%%%%%%%%%%%%%%%%%%%%%%%%%%%%
% APPENDIX
%%%%%%%%%%%%%%%%%%%%%%%%%%%%%%%%%%%%%%%%%%%%%%%%%%%%%%%%%%%%%%%%%%%%%%%%%%%%%%%
%%%%%%%%%%%%%%%%%%%%%%%%%%%%%%%%%%%%%%%%%%%%%%%%%%%%%%%%%%%%%%%%%%%%%%%%%%%%%%%
% \newpage
\appendix
%\onecolumn
% \section{Appenix A: Theory}
%%%%%%%%%%%%%%%%%%%%%%%%%%%%%%%%%%%%%%%%%%%%%%%%%%%%%%%%%%%%%%%%%%%%%%%%%%%%%%%
%%%%%%%%%%%%%%%%%%%%%%%%%%%%%%%%%%%%%%%%%%%%%%%%%%%%%%%%%%%%%%%%%%%%%%%%%%%%%%%

\section{Proof: Optimiser Is Controller}

\subsection{SGD Is a P Controller}

The parameter update rule of SGD from iteration $t$ to $t+1$ is determined by
\begin{equation} \label{Eq21}
\theta_{t+1}=\theta_t-r \partial L_t / \partial \theta_t
\end{equation}

\noindent where $r$ is the learning rate. We now regard the gradient $\partial L_t / \partial \theta_t$ as error $e(t)$ in the PID control system \cite{wang2020pid}. Compared to the PID controller, we find that SGD can be viewed as one type of $P$ controller with $K_p=r$. The system function of SGD becomes:

\begin{equation} \label{Eq22}
\theta_{SGD}(s)=r
\end{equation}

\subsection{SGDM Is a PI Controller}

SGDM, which leverages historical gradients, trains a DNN more swiftly than SGD does. The rule of SGDM updating parameter is given by
\begin{equation} \label{Eq23}
\left\{\begin{array}{l}
V_{t+1}=\alpha V_t-r \partial L_t / \partial \theta_t \\
\theta_{t+1}=\theta_t+V_{t+1}
\end{array}\right.
\end{equation}

\noindent where $V_t$ is a term that accumulates historical gradients. $\alpha \in(0,1)$ is the factor that balances the past and current gradients. It is usually set to $0.9$. Dividing two sides of the Equation \ref{Eq23} by $\alpha^{t+1}$, we get: 
\begin{equation} \label{Eq24}
\frac{V_{t+1}}{\alpha^{t+1}}=\frac{V_t}{\alpha^t}-r \frac{\partial L_t / \partial \theta_t}{\alpha^{t+1}} .
\end{equation}

\noindent Finally, we get $\theta_{t+1}$ as follow by iteration:
\begin{equation} \label{Eq25}
\theta_{t+1}-\theta_t=-r \frac{\partial L_t}{\partial \theta_t}-r \sum_{i=0}^{t-1} \alpha^{t-i} \frac{\partial L_i}{\partial \theta_i}
\end{equation}

\noindent SGDM actually is a PI controller with $K_p=r$ and $K_i=r\alpha^{t-i}$. The system function of SGDM should be:

\begin{equation} \label{Eq26}
\theta_{SGDM}(s)=r + \frac{r}{s} \cdot \frac{1}{s-ln(\alpha)} 
\end{equation}

\subsection{PID optimiser Is a PID Controller}

SGD and SGDM can be respectively viewed as P and PI controller \cite{wang2020pid}. Given that training is often conducted in a mini-batch manner, the learning process is very easy to introduce noise when computing gradients. The proposed PID optimiser \cite{wang2020pid} updates network parameter $\theta$ in iteration $(t+1)$ by
\begin{equation} \label{Eq27}
\left\{\begin{array}{l}
V_{t+1}=\alpha V_t-r \partial L_t / \partial \theta_t \\
D_{t+1}=\alpha D_t+(1-\alpha)\left(\partial L_t / \partial \theta_t-\partial L_{t-1} / \partial \theta_{t-1}\right) \\
\theta_{t+1}=\theta_t+V_{t+1}+K_d D_{t+1} .
\end{array}\right.
\end{equation}

\noindent Thus, the $\theta_{t+1}$ using PID optimiser is described as follow by iteration:
\begin{equation} \label{Eq28}
\theta_{t+1}-\theta_t=-r \frac{\partial L_t}{\partial \theta_t} - r \sum_{i=0}^{t-1} \alpha^{t-i} \frac{\partial L_i}{\partial \theta_i} - r K_{d} \left( \frac{\partial L_i}{\partial \theta_i} - \frac{\partial L_{i-1}}{\partial \theta_{i-1}} \right)
\end{equation}

\noindent where $K_{d} \left( \frac{\partial L_i}{\partial \theta_i} - \frac{\partial L_{i-1}}{\partial \theta_{i-1}} \right)$ is the D component of the PID controller. The system function of PID should be:

\begin{equation} \label{Eq29}
\theta_{PID}(s) = r + \frac{r}{s} \cdot \frac{1}{s-ln(\alpha)} + K_{d} s
\end{equation}

When setting the hyperparameter $\alpha=1.0$, we can get the vanilla PID optimiser: $K_{p}=r$, $K_{i}=r$ and $K_{d}=r \cdot K_{d}$

\subsection{AdaM Is a PI Controller with an Adaptive Filter}

Based on adaptive estimates of lower-order moments, AdaM algorithm adaptively adjusts the stochastic gradients, and it can be summarized as below:

\begin{equation} \label{Eq30}
\left\{\begin{array}{l}
m_{t+1}=\beta_{1} m_t + \left(1-\beta_{1}\right) \partial L_t / \partial \theta_t \\
v_{t+1}=\beta_{2} v_t + \left(1-\beta_{1}\right) \partial L_t / \partial \theta_t \\
\hat{m}_{t+1}=m_{t} / \left( 1 - \beta^{t}_{1} \right) \\
\hat{v}_{t+1}=v_{t} / \left( 1 - \beta^{t}_{2} \right) \\
\theta_{t+1}=\theta_t + \alpha \hat{m}_{t} / \left( \sqrt{\hat{v}_{t+1}} + \epsilon \right)
\end{array}\right.
\end{equation}

where $m_{t}$ is the first moment estimate at timestep $t$, and $v_{t}$ is the second raw moment estimate. The default set of learning rate $\alpha$, hyperparameters $\beta_{1}$, $\beta_{2}$ and $\epsilon$ are respectively $0.001$, $0.9$, $0.999$ and $10^{ - 8}$.

\noindent The iteration of $\theta_{t+1}$ using the AdaM optimizer is described as follows:

\begin{equation} \label{Eq31}
\begin{aligned}
\theta_{t+1} & = \theta_{t} - r \cdot \frac{\widehat{m}_{t}}{\sqrt{\widehat{v}_{t}} + \epsilon} \\
& = \theta_{t} - r \cdot \frac{\frac{\sum_{i=0}^{t}\beta_{1}^{t-i}(\partial L_{i} / \partial \theta_{i})}{ \sum_{i=1}^{t}\beta_1^{i-1} }}{\sqrt{{\frac{\sum_{i=1}^{t}\beta_{2}^{t-i}(\partial L_{i} / \partial \theta_{i})^2}{ \sum_{i=1}^{t}\beta_2^{i-1} }}} + \epsilon} \\
& = \theta_{t} - r \cdot \frac{1}{M} \beta_{1}^{0} \frac{\partial L_{t}}{\partial \theta_{t}} - r \cdot \frac{1}{M} \sum_{i=0}^{t-1}\beta_{1}^{t-1-i} \frac{\partial L_{i}}{\partial \theta_{i}}
\end{aligned}
\end{equation}

where $M$ is the adaptive part of AdaM, and its formula is:

\begin{equation} \label{Eq32}
M = \frac{1}{\sqrt{\frac{ \sum_{i=0}^{t}\beta _{2}^{t-i}( \partial L_t / \partial \theta_t)^{2} }{ \sum_{i=0}^{t}\beta_{2}^{i-1} }} + \epsilon } \cdot \frac{ 1 }{ \sum_{i=0}^{t} \beta_{1}^{i-1} }
\end{equation}

Compared to Equation \ref{Eq25}, the adaptive component $M$ of AdaM plays an important role on adapting the learning system. We cannot derive the system function of AdaM, as the high complexity of $M$. Finally, we directly use the same S function in SIMULINK and get its system response on above mentioned ANN models.

\subsection{Filter Processed SGD}

Although Gaussian LPF-SGD outperforms other SGD variants, we still do not know which part it has filtered, for example, high frequency, low frequency or any band frequency parts. In this study, we summarize the SGD learning process under the processing of filters as below:

\begin{equation} \label{Eq33}
\left\{\begin{array}{l}
\widehat{\partial L_t / \partial \theta_t} = \partial L_t / \partial \theta_t + \partial \left( \int_{-\infty}^{\infty} L(\theta_{t}-\tau) H(\tau) d \tau \right) / \partial \theta_t\\
\theta_{t+1} = \theta_t - r \widehat{\partial L_t / \partial \theta_t}
\end{array}\right.
\end{equation}

where $H$ is a Gaussian kernel and $L(\theta_{t})$ is the loss function of the training process in GLPF-SGD \cite{bisla2022low}. The $\theta_{t+1}$ using $Filter$ processed SGD optimiser is described as follow by iteration:

\begin{equation} \label{Eq34}
\begin{aligned}
\theta_{t+1} & = \theta_t - r \frac{\partial L_t}{\partial \theta_t} + r \frac{\partial \left( \int_{-\infty}^{\infty} L(\theta_{t}-\tau) H(\tau) d \tau \right)}{\partial \theta_t}\\
& = \theta_{t} - r \frac{\partial L_{t}}{\partial \theta_{t}} + r \frac{\partial \left( L {\circledast} H \right)}{\partial \theta_{t}}\\
& = \theta_t - r \frac{\partial L_t}{\partial \theta_t} + r \frac{1}{G}\frac{\sum_{i=0}^N \partial \left( L(\theta_t-\tau_i) H(\tau_i)\right)}{\partial \theta_t} 
\end{aligned}
\end{equation}

where $G$ is the gain of the filter $H$ with the order of $N$, and $\circledast$ is the convolution process. Finally, the system function of filter processed SGD becomes to:

\begin{equation} \label{Eq35}
\theta_{FP-SGD}(s) = r \left(Gain \cdot \frac{\prod_{i=0}^{m} \left(s+h_{i}\right)}{\prod_{j=0}^{n} \left(s+l_{j}\right)} \right)
\end{equation}

% \theta_{t+1} & = \theta_{t} - r \cdot \frac{\widehat{m}_{t}}{\sqrt{\widehat{v}_{t}} + \epsilon} \\
% & = \theta_{t} - r \cdot \frac{\frac{\sum_{i=0}^{t}\beta_{1}^{t-i}(\partial L_{i} / \partial \theta_{i})}{ \sum_{i=1}^{t}\beta_1^{i-1} }}{\sqrt{{\frac{\sum_{i=1}^{t}\beta_{2}^{t-i}(\partial L_{i} / \partial \theta_{i})^2}{ \sum_{i=1}^{t}\beta_2^{i-1} }}} + \epsilon} \\
% & = \theta_{t} - r \cdot \frac{1}{M} \beta_{1}^{0} \frac{\partial L_{t}}{\partial \theta_{t}} - r \cdot \frac{1}{M} \sum_{i=0}^{t-1}\beta_{1}^{t-1-i} \frac{\partial L_{i}}{\partial \theta_{i}}\\

In this study, to analyse which frequency parts are beneficial to the training, we used a second-order Infinite Impulse Response (IIR) filter instead of the Gaussian kernel filter. By approximately setting the cutoff frequency at half, we imply a low-pass filter ranging from 0 Hz to half the sampling rate and a high-pass filter from half the sampling rate up to the sampling rate. Consequently, knowledge of the exact sampling rate is unnecessary, and essentially, it remains unobtainable.

\section{Proof: Learning Systems of Most ANNs are Control Systems}

\subsection{CNN and Its Control System}

\begin{wrapfigure}{r}{0.5\textwidth}
\centering
\includegraphics[scale=0.34]{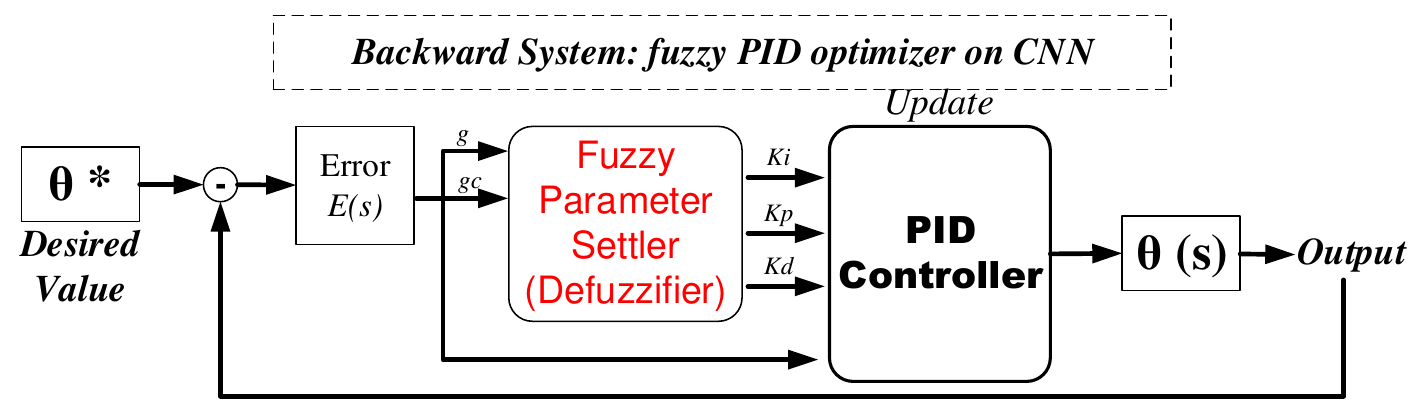}
\captionof{figure}{The control system of CNN updated by the FuzzyPID optimiser.}
\label{Figure6}
\end{wrapfigure}

Most CNNs have been used to perform the classification task using the backpropagation algorithm. Obviously, this learning system is a single-input-single-output (SISO) control system, indicating that each sample corresponds to a single label. Figure \ref{Figure6} provides a concise representation of the learning structure when focusing solely on the optimizer, exemplified here by the fuzzyPID optimizer applied to CNNs. If considering only optimiser, its brief learning structure can be seen in Figure \ref{Figure6}, and we give an example of using fuzzyPID optimiser on CNNs. When using each $Controller$, the Laplace transform of the learning process becomes:

\begin{equation} \label{Eq36}
\begin{aligned}
\theta(s) & = Controller \cdot E(s)\\
& =Controller \cdot \left( \frac{\theta^{\ast}}{s}-\theta(s)\right)\\
& = \frac{Controller}{Controller+1} \cdot \frac{\theta^{\ast}}{s}\\
\end{aligned}
\end{equation}

Backpropagation algorithm based ANN models rely on the backward error to update weights themselves, and inevitably, the system function of their learning systems have been determined by such designed algorithm. Therefore, there are two factors can significantly affect their performance. One is hyperparameter that setups high techniques on ANN models, and another one is optimiser that controls the convergence speed and stability. 

It is clear that the network parameter update using SGD optimiser depends on current gradient $r \partial L_t / \partial \theta_t$ , but other well-performed updating methods, such as SGDM, AdaM and PID, have considered the previous gradient. The accumulation part of gradients in SGDM can accelerate the learning process, and the introduction of decay term $\alpha$ is to keep the gradients away from the current value so that it can alleviate noise. Building on SGDM, the PID optimizer introduces the predicted future trend (the difference between the current gradient and the previous one) to adjust the learning process, and its speed becomes faster than SGDM. However, coefficients of PID optimiser, such as P, I and D, are totally fixed, and that will bring another problem -- overshooting. To counteract this issue, we used fuzzy logic to adaptively adjust the coefficients of PID optimiser. Inspired by the GLPF-SGD, we believe the learning process using any optimiser relies on specific frequency components. In this study, we designed two filters to figure out which frequency component ANN models prefer. To avoid a long lag of convolution computing, we only applied a second-order IIR filter on the SGD learning process. Even without the exact sampling rate, we have chosen from the half, as the frequency component has no relationship with the sampling rate if we cutoff from the $2^{-1*i}$ of sampling rate. Therefore, we can get determined $\theta(s)$ of backpropagation based learning systems using various optimisers as follow:

\noindent \textbf{(1)} When $Controller=\theta_{SGD}(s)$, we can get $\theta(s)$ of backpropagation based CNNs as below:
\begin{equation} \label{Eq37}
\theta(s) = \frac{K_{p}}{K_{p}+1} \cdot \frac{\theta^\ast}{s}
\end{equation}

\noindent \textbf{(2)} When $Controller=\theta_{SGDM}(s)$, and if we set $\alpha=1.0$, we can get $\theta(s)$ of backpropogation based CNNs using SGDM as the optimiser as below:

\begin{equation} \label{Eq38}
\theta(s) = \frac{K_{p}s+K_{i}}{(K_{p}+1)s+K_{i}} \cdot \frac{\theta^\ast}{s} 
\end{equation}

\noindent \textbf{(3)} Based on prior knowledge of control system engineering, PID optimiser was proposed by adding D component on SGDM optimiser. According to the analysis of PID optimiser \cite{wang2020pid} and the Ziegler–Nichols optimum setting rule \cite{ziegler1942optimum}, we also set $P=1$, $I=5$ and $D=100$ here. Therefore, when $Controller=\theta_{PID}(s)$, $\theta(s)$ can be computed by:
\begin{equation} \label{Eq39}
\begin{aligned}
\theta(s) & = \frac{K_{p}+K_{i}\frac{1}{s}+K_{d}s}{K_{p}+K_{i}\frac{1}{s}+K_{d}s+1} \cdot \frac{\theta^\ast}{s}\\
& = \frac{K_{d}s^2+K_{p}s+K_{i}}{K_{d}s^2+(K_{p}+1)s+K_{i}} \cdot \frac{\theta^\ast}{s}
\end{aligned}
\end{equation}

\noindent \textbf{(4)} Considering the use of fuzzy logic on PID optimiser, we finally get Equation \ref{Eq13} that can compute the system response of FuzzyPID on backpropogation based ANNs. 

\noindent \textbf{(5)} When using AdaM as the optmiser, apart from the adaptive part, we found AdaM shares the same parameter updating strategy as the SGDM. With $\beta_1=0.9$, the Laplace transform of AdaM becomes:

\begin{equation} \label{Eq40}
\begin{aligned}
AdaM(s) & = \frac{1}{M} \cdot K_{p} + \frac{1}{M} \cdot K_{i} \frac{1}{s} \cdot \frac{1}{s-ln(\beta_1)}\\
\end{aligned}
\end{equation}

where the Laplace transform of the last term in Equation \ref{Eq28} becomes to:

\begin{equation} \label{Eq41}
\begin{aligned}
\mathnormal{Laplace} \left( \sum_{i=0}^{t-1}\beta_{1}^{t-1-i} \frac{\partial L_{i}}{\partial \theta_{i}} \right) & = \mathnormal{Laplace} \left( \sum_{i=0}^{t-1}e^{ln\beta_{1}(t-1-i)} \right) \cdot \mathnormal{Laplace} \left( \sum_{i=0}^{t-1} \frac{\partial L_{i}}{\partial \theta_{i}} \right)\\
& = \frac{1}{s-ln(\beta_1)} \cdot \frac{1}{s}\\
\end{aligned}
\end{equation}

Hence, the system function $\theta(s)$ when using AdaM as the optimiser is:

\begin{equation} \label{Eq42}
\begin{aligned}
\theta(s) & = \frac{AdaM(s)}{AdaM(s) + 1} \cdot \frac{\theta^\ast}{s}\\
& = \frac{K_{p}s + K_{i}} {Ms^2 + (K_{p}-Mln\beta_{1})s+K_{i}} \cdot \frac{\theta^\ast}{s}\\
\end{aligned}
\end{equation}

\noindent \textbf{(5)} Additionally, when $Controller=\theta_{FP-SGD}(s)$ the system function of using SGD processed with $Filter$ is defined as:

\begin{equation} \label{Eq43}
\begin{aligned}
\theta(s) & = \frac{ Filter}{Filter + 1} \cdot \frac{\theta^\ast}{s}\\
 & = \frac{Gain \cdot \prod_{i=0}^{m} \left(s+h_{i}\right)}{Gain \cdot \prod_{i=0}^{m} \left(s+h_{i}\right) + \prod_{j=0}^{n} \left(s+l_{j}\right)} \cdot \frac{\theta^\ast}{s}\\
\end{aligned}
\end{equation}

\subsection{FFNN and Its Control System}

\begin{wrapfigure}{r}{0.5\textwidth}
\centering
\includegraphics[scale=0.4]{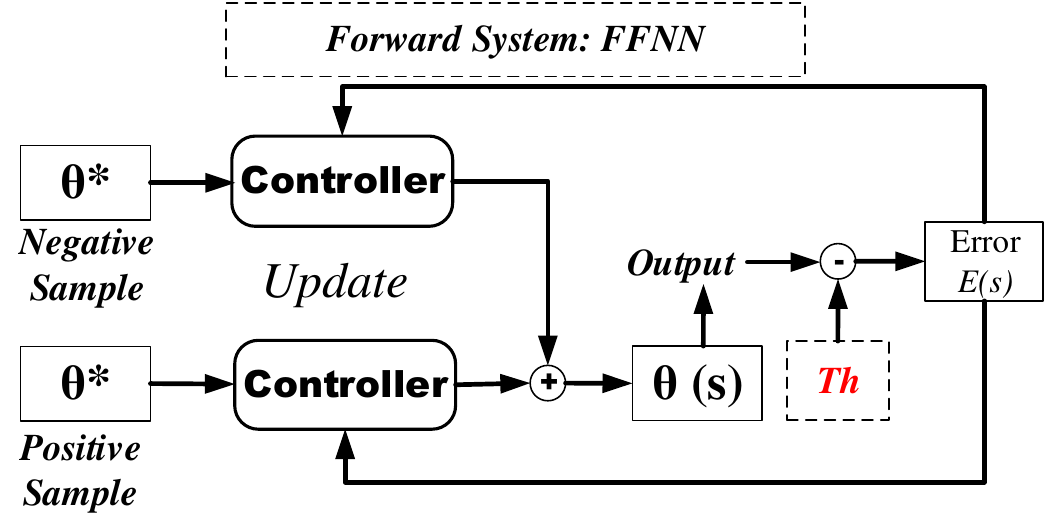}
\captionof{figure}{The control system of FFNN.}
\label{Figure7}
\end{wrapfigure}

FFNN \cite{hinton2022forward}, based on the forward-forward algorithm mainly aims to visualize the learning process. For a clear analysis on FFNN, we set the portion of positive samples $\lambda=0.5$ and the threshold $Th=1.0$. These two hyperparameters were used to make the goodness be well above some threshold value for real data and well below that value for negative data. Essentially, we still use the backpropagation algorithm to update the weights for each layer, as one method used in \cite{hinton2022forward}. Based on Equation \ref{Eq14} and \ref{Eq15}, we found when $\lambda=0.5$, the optimal result $\theta^{\ast}$ has no relationship with the learning system. We analysed FFNN on seven optimisers (e.g., SGD, SGDM, AdaM , PID, LPF-SGD, HPF-SGD and FuzzyPID) and derived their control system functions as below:

\begin{equation} \label{Eq44}
\begin{aligned}
\theta(s) & =  \left( -(1 -\lambda)\frac{\theta^{\ast}}{s} + \lambda\frac{\theta^{\ast}}{s}-\left[ \theta(s)-\frac{Th}{s}\right] \right) \cdot Controller\\
& = \frac{1}{Controller+1} \cdot \left( \frac{(2\lambda-1)\theta^{\ast}+Th}{s} \right) 
\end{aligned}
\end{equation}

But when $\lambda \neq 0.5$ and the threshold $Th \neq 1.0$, the system function and its classification performance will be influenced by these two hyperparameters. The product of $(2\lambda-1)\theta^{\ast}+Th$ is a gain adjustment part that will affect the learning process.

\subsection{GAN and Its Control System}

The essence of GAN is that G and D play games with each other and finally reach a Nash equilibrium point \cite{kreps1989nash}, but this is only an ideal situation. The normal situation is that it is easy for one party to be strong and the other party to be weak. Therefore, two problems appeared \textbf{(1)} Gradient disappearance and \textbf{(2)} mode collapse corresponding to D and G being the result of the stronger side.

\begin{wrapfigure}{r}{0.5\textwidth}
\centering
\includegraphics[scale=0.38]{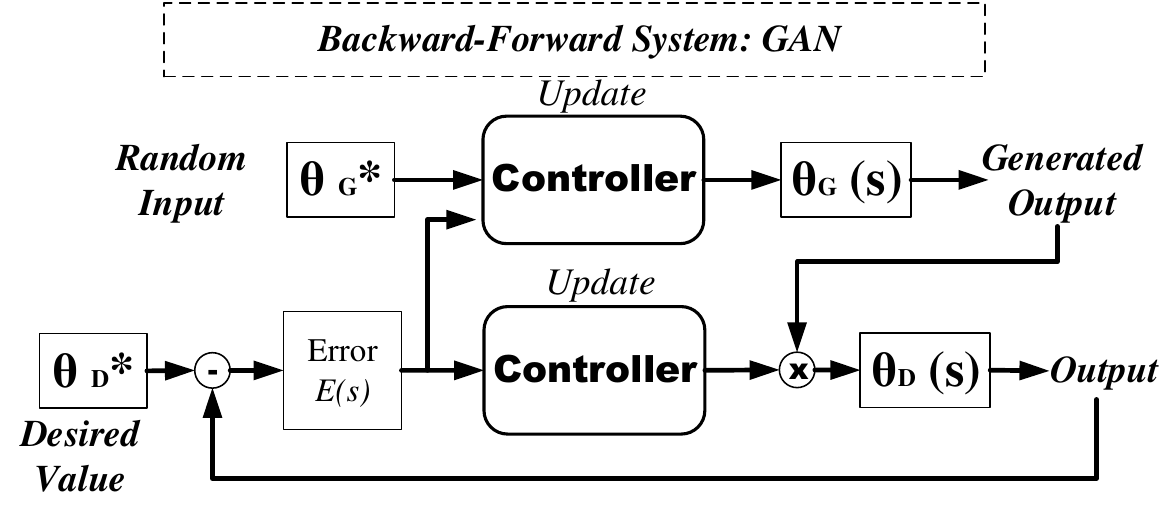}
\captionof{figure}{The control system of classical GAN.}
\label{Figure8}
\end{wrapfigure}

The situation of gradient vanishing is that D wins the game. Because the gradient update of G comes from D, and in the initial stage of training, the input of G is randomly generated noise, which will definitely not generate good pictures, but D performs well. It is easy to judge the true and false samples, that is, there is almost no loss in the training of D. Therefore, there is no effective gradient information back to G itself. 

The problem of mode collapse is mainly that G wins the game. That causes D to be unable to distinguish between real pictures and fake pictures generated by G. But D cannot tell the difference, and give the correct evaluation, then G will think that this picture is correct. Thus, D still gives the correct evaluation. Therefore, these two ANNs are such mutual deception.

For a clear analysis on GAN, we used a classical GAN model \cite{2014Generative}. We derive its control system function (Seen from Figure \ref{Figure8}) under seven optimisers:

\textbf{(1)} When using SGD as the optimiser, $Controller=K_{p}$ we get the the control system of G as below:
\begin{equation} \label{Eq45}
\begin{aligned}
\theta_G(s) & = \frac{1}{2} \cdot \left( \frac{\theta_{D}^{\ast}}{K_{p}} \pm
\sqrt{ \left(\frac{{\theta_D^\ast}}{K_{p}} \right)^{2} - \frac{4}{s}} \right)
\end{aligned}
\end{equation}

\textbf{(2)} When using SGDM (a PI controller) as the optimiser, $controller=K_{p}+K_{i}/s$ , we get the the control system of G as below:
\begin{equation} \label{Eq46}
\theta_{G}(s) = \frac{1}{2} \cdot \left( \frac{\theta_{D}^{\ast}s}{K_{p}s+K_{i}} \pm \sqrt{\left( \frac{\theta_{D}^{\ast}}{K_{p}s+K_{i}} \right) ^{2} - \frac{4}{s}} \right)
\end{equation}

\textbf{(3)} When using AdaM (merging the PI and an adaptive filter) as the optimiser, $controller=AdaM(s))$, we get the the control system of G as below:
\begin{equation} \label{Eq47}
\theta_G(s) = \frac{1}{2} \cdot \left( \frac{\theta_{D}^{\ast}}{AdaM(s)} \pm \sqrt{ \left(\frac{\theta_{D}^\ast}{AdaM(s)} \right)^{2} - \frac{4}{s}} \right)
\end{equation}

\textbf{(4)} When using PID (considering the pass, current and future) as the optimiser, $controller=K_{p}+K_{i}/s+K_{d}s$. Finally, we get the control system of G as below:
\begin{equation} \label{Eq48}
\theta_G(s) = \frac{1}{2} \cdot \left( \frac{\theta_{D}^{\ast}}{K_{p}+\frac{K_{i}}{s}+K_{d}s} \pm \sqrt{ \left(\frac{\theta_D^\ast}{K_{p}+\frac{K_{i}}{s}+K_{d}s} \right)^{2} - \frac{4}{s}} \right)
\end{equation}

\textbf{(5)} When using $Filter$ processed SGD as the optimiser, $controller=Filter$, we get the the  control system of G as below:
\begin{equation} \label{Eq49}
\theta_G(s) = \frac{1}{2} \cdot \left( \frac{\theta_{D}^{\ast}}{Gain \cdot \frac{\prod_{i=0}^{m} \left(s+h_{i}\right)}{\prod_{j=0}^{n} \left(s+l_{j}\right)}} \pm \sqrt{ \left(\frac{\theta_D^\ast}{Gain \cdot \frac{\prod_{i=0}^{m} \left(s+h_{i}\right)}{\prod_{j=0}^{n} \left(s+l_{j}\right)}} \right)^{2} - \frac{4}{s}} \right)
\end{equation}

Owing to the complexity of their system functions, we finally decided to use MATLAB SIMULINK to analyse their system response and stability, as shown in Figure \ref{Figure5}.

\section{Residual Connections}

\begin{figure}[h]
    \centering 
    \includegraphics[scale=0.38]{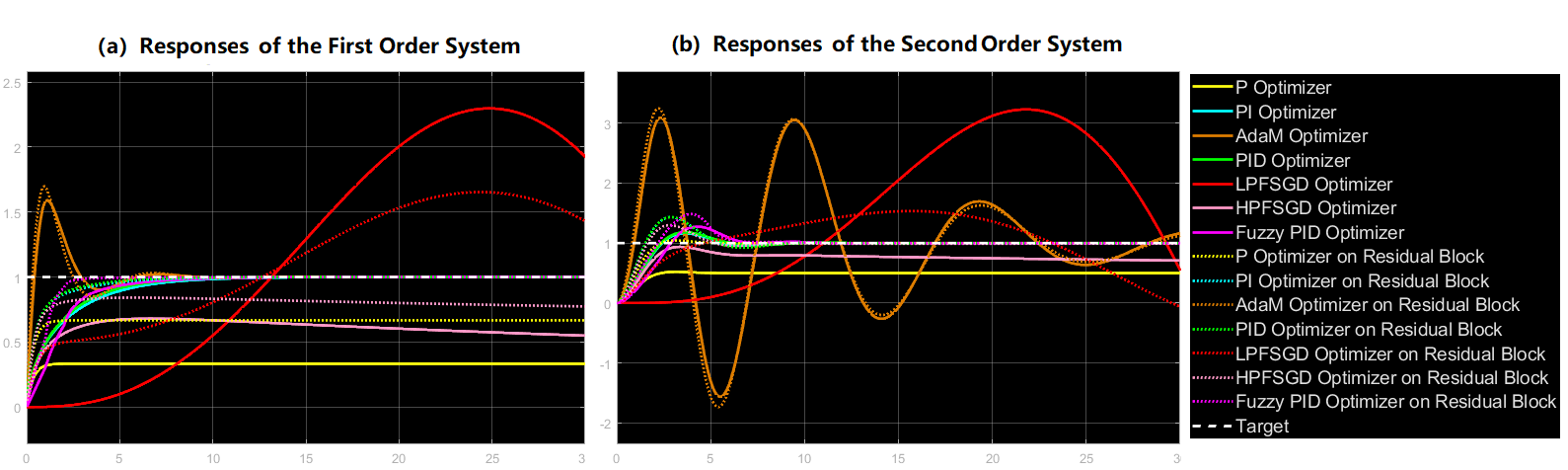} 
    \caption{Step responses of models with residual connections across various optimizers: such as SGD, SGDM, AdaM, PID, LPF-SGD, HPF-SGD and FuzzyPID optimisers.} 
    \label{Figure10} 
\end{figure}

\begin{wrapfigure}{r}{0.5\textwidth}
\centering
\includegraphics[scale=0.34]{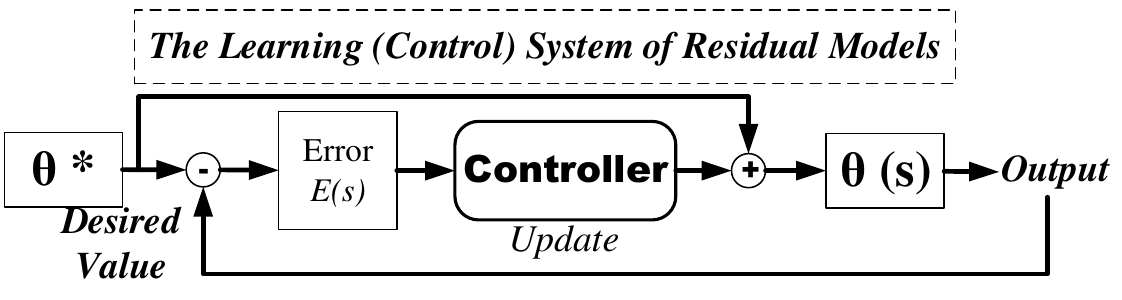}
\captionof{figure}{Control system of Residual models.}
\label{Figure9}
\end{wrapfigure}

Residual connections (RSs) \cite{he2016identity,eunice2022deep} aim to ease the training of DNNs, and it can \textbf{(1)} increase the depth of ANNs and \textbf{(2)} avoid gradient vanishing. Most state-of-the-art (SOTA) ANN models have RSs, but the use of such forward connections has no systematic analysis. Compared to a plain CNN layer, Residual Block adds a short cut from the input features to the output of the mapping. The output from the residual block is $H(x) = f(x) + x$, where input features are $x$, the output from the original mapping is $f(x)$ , then, our desired output is $H(x)$. We are learning the residuals from the output in relate to the input, as the RS is trying to fit the mapping $f(x) = H(x) - x$.

In this study, the models we designed with two or four hidden layers constitute a first-order system, and their system response can also be seen in Figure \ref{Figure10}(a). However, we assume SOTA models, such as VGG19 \cite{simonyan2014very}, ResNet18, ResNet50, ResNet101 \cite{he2016deep}, DenseNet121 \cite{zhu2017densenet}, MobileNetV2 \cite{sandler2018mobilenetv2}, EffecientNet \cite{tan2019efficientnet}, are second-order (or higher) models. According to the computing process of SGD and the explanation of ResNet \cite{he2016identity}, we present a reformulated RS mechanism below (ignoring BN \cite{ioffe2015batch}, ReLU \cite{nair2010rectified}, pooling \cite{wu2015max}, and exponential or cosine decay \cite{li2021second}):

\begin{equation} \label{Eq50}
\frac{\partial L_t}{\partial \hat{\theta}_t}\frac{\partial \hat{\theta}_t}{\partial \theta_t} = \frac{\partial L_t}{\partial \hat{\theta}_t} \left(1+\frac{\partial}{\partial \theta_t} \sum_{i=l}^{L-1} \mathcal{F}\left(\theta_i\right)\right)
\end{equation}

where $\mathcal{F}$ is a residual function, $\hat{\theta}_t$ is the weights in the residual block, and $L$ (also interpretable as the depth of a single residual block) is the deeper unit in \cite{he2016identity}. Equation \ref{Eq50} indicates that the gradient $\partial L_t / \partial \theta_t$ can be decomposed into two additive terms: a term of $\partial L_t / \partial \theta_t$ that propagates information directly without concerning any weight layers, and another term of $\frac{\partial}{\partial \theta_t} \sum_{i=l}^{L-1} \mathcal{F}\left(\theta_i\right)$ that propagates through the weight layers. The additive term of (or this direct component) $\partial L_t / \partial \theta_t$ determines that the learning system will consider information which directly propagates back to $\theta_t$. The parameter update rule of SGD from iteration $t$ to $t + 1$ using RSs is determined by : 

\begin{equation} \label{Eq51}
\theta_{t+1} = \theta_t - r \partial L_t / \partial \theta_t - r
\end{equation}

where we assume the residual block $ \mathcal{F} $ is a simple block. Thus, we finally get the system function of residual connections $\theta_{RS}(s)$ as below:

\begin{equation} \label{Eq52}
\begin{aligned}
\theta_{RS}(s) & = r+\frac{r}{s}\\
\end{aligned}
\end{equation}

where the learning rate $r$ can be served as $K_{p}$, and $\frac{r}{s}$ is aligns with the second part $\frac{K_{i}\alpha^{t-i}}{s}$ of SGDM in Equation \ref{Eq25}. The difference is that SGDM has a momentum that takes previous gradients into account, but RS integrates information from preceding layers. Analysing a node within RS-based ANN models, we found the system function of these two -- SGDM and RS -- have a very similar format. SGDM optimizes the weight of models by accumulating previous gradients with the use of a momentum factor to adjust the effect of accumulation on the time dimension. However, RS optimizes the model by adding passed information to the current block on the space dimension.

In Figure \ref{Figure14}, models with residual connections, such as ResNet50, DenseNet121, ModelNetV2, and EffecientNet, have the classification advantage using SGDM and PID, even though PID displays irregularities in the training curve. Interestingly, this observation is also echoed in Figure \ref{Figure10}(b), as the rising time of SGDM and PID controller on residual connections is shorter than others (except AdaM, although AdaM can rise very fast, it demonstrates heightened oscillations). FuzzyPID trails closely, while LPF-SGD lags due to its pronounced low-frequency characteristics leading to the most gradual climb.

\section{CycleGAN}

\begin{figure}[h]
    \centering 
    \begin{subfigure}[b]{0.24\textwidth} 
        \centering 
        \includegraphics[scale=0.22]{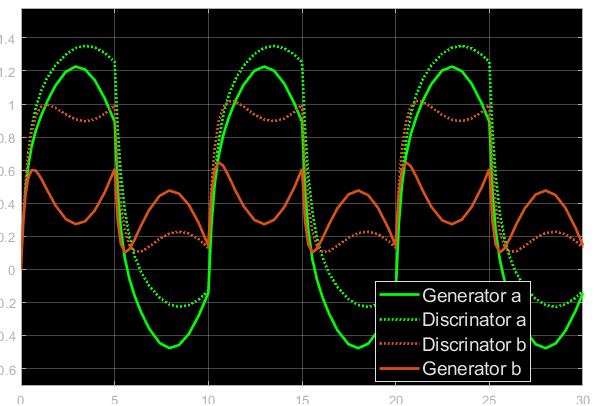} 
        \caption{\small CycleGAN on SGD.} 
        \label{Figure12-a} 
    \end{subfigure} 
    \begin{subfigure}[b]{0.24\textwidth}
        \centering 
        \includegraphics[scale=0.22]{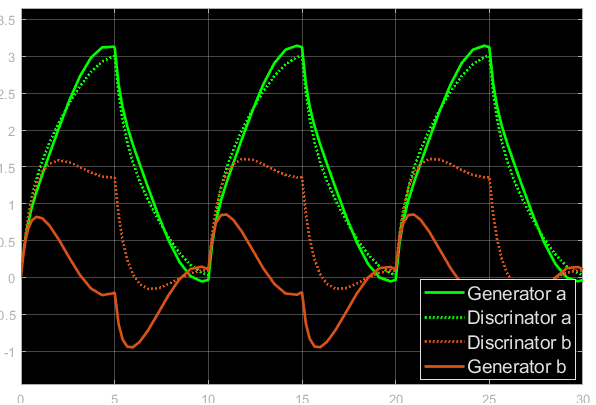} 
        \caption{\small CycleGAN on SGDM.} 
        \label{Figure12-b} 
    \end{subfigure}
    \begin{subfigure}[b]{0.24\textwidth}
        \centering 
        \includegraphics[scale=0.22]{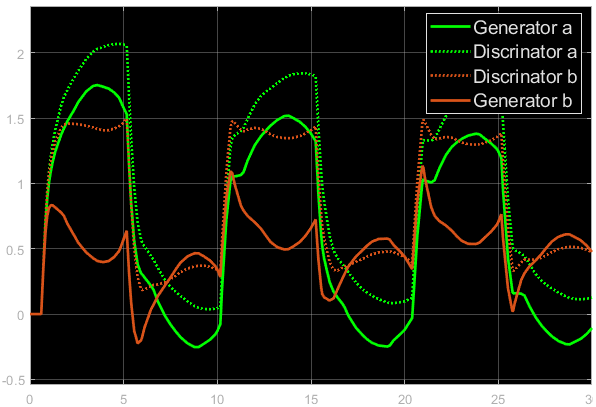} 
        \caption{\small  CycleGAN on AdaM.} 
        \label{Figure12-c} 
    \end{subfigure}
    \begin{subfigure}[b]{0.24\textwidth}
        \centering 
        \includegraphics[scale=0.22]{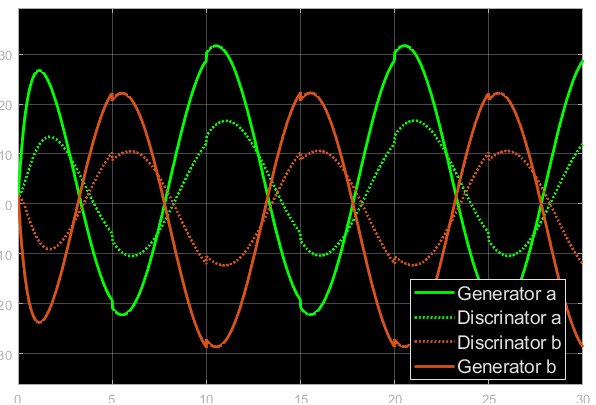} 
        \caption{\small  CycleGAN on PID.} 
        \label{Figure12-d} 
    \end{subfigure}
    \begin{subfigure}[b]{0.24\textwidth}
        \centering 
        \includegraphics[scale=0.22]{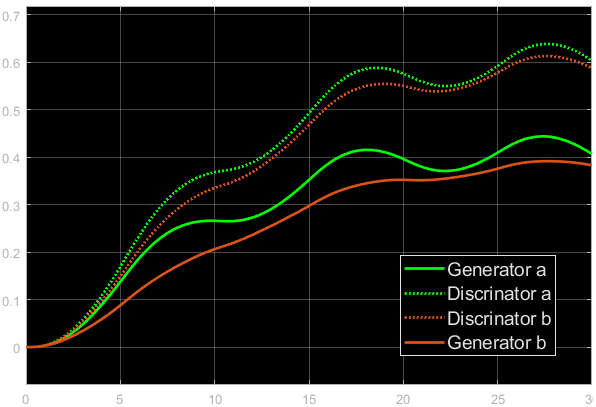} 
        \caption{\small  CycleGAN on LPF-SGD.} 
        \label{Figure12-e} 
    \end{subfigure}
    \begin{subfigure}[b]{0.24\textwidth}
        \centering 
        \includegraphics[scale=0.22]{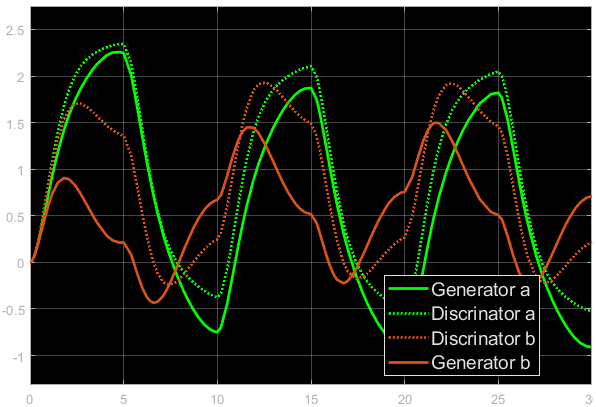} 
        \caption{\small  CycleGAN on HPF-SGD.} 
        \label{Figure12-f} 
    \end{subfigure}
        \begin{subfigure}[b]{0.24\textwidth}
        \centering 
        \includegraphics[scale=0.22]{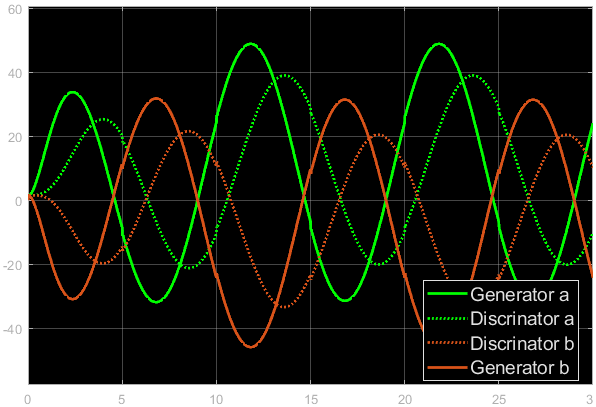} 
        \caption{\small  CycleGAN on FuzzyPID.} 
        \label{Figure12-g} 
    \end{subfigure}
    \caption{The system response of CycleGAN on different hyperparameters and optimisers, such as SGD, SGDM, AdaM, PID, LPF-SGD, HPF-SGD and FuzzyPID optimisers.} 
    \label{Figure12} 
\end{figure}

\begin{wrapfigure}{r}{0.54\textwidth}
\centering
\includegraphics[scale=0.34]{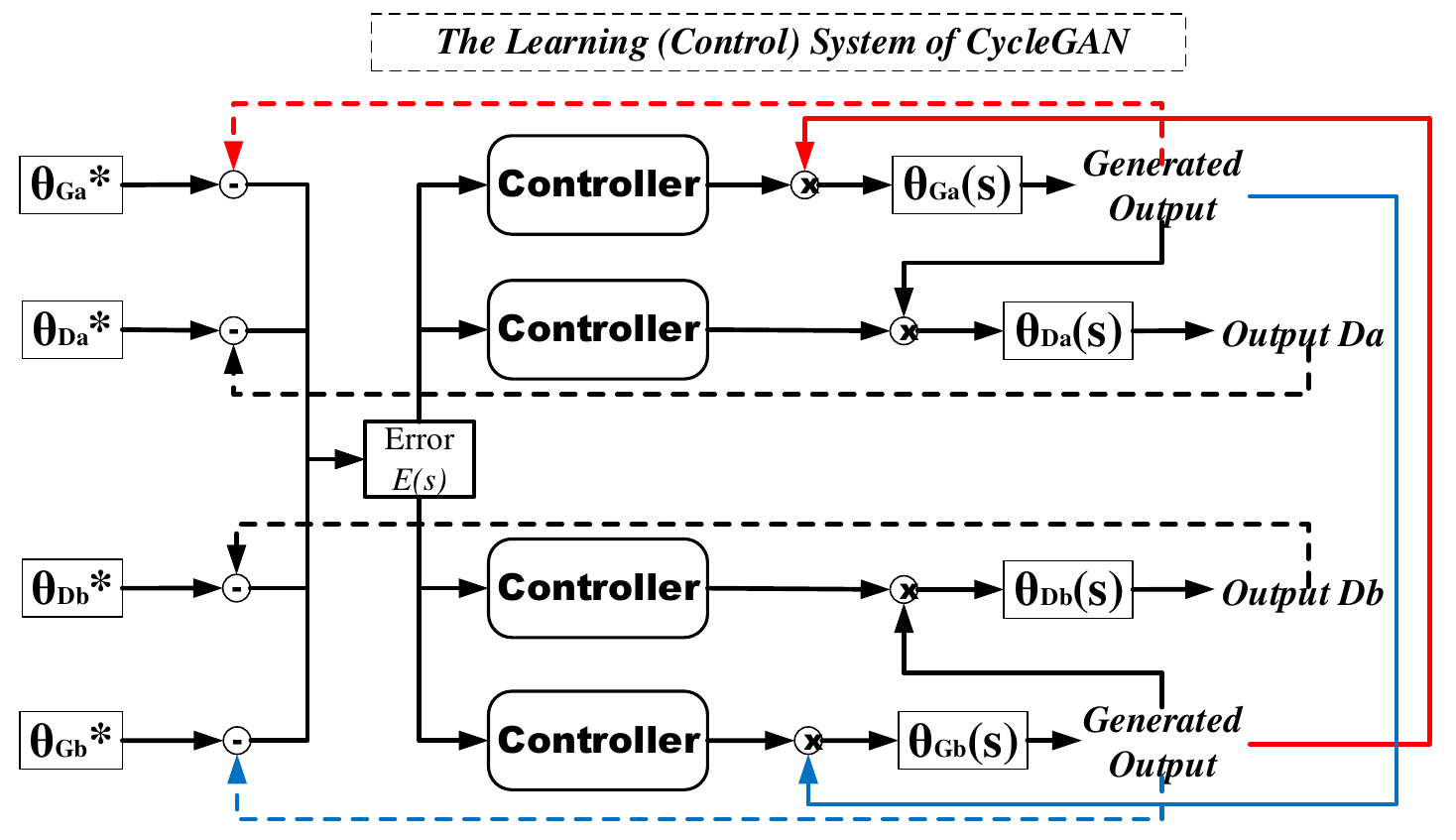}
\captionof{figure}{The control system of CycleGAN.}
\label{Figure11}
\end{wrapfigure}

CycleGAN \cite{zhu2017unpaired} aims to translate an image from a source domain $a$ to a target domain $b$ in the absence of paired examples. We denote the data distribution as $a \thicksim p_{data}(a)$ and $b \thicksim p_{data}(b)$. CycleGAN contains two mapping functions $G_{a}$: $A \rightarrow B$ and $G_{b}$: $B \rightarrow A$, and associated adversarial discriminators $D_{a}$ and $D_{b}$. {$D_{b}$ encourages generator $G_{a}$ to translate $A$ into outputs indistinguishable from domain $B$, and vice versa for $D_{A}$ and $B$. According to its learning system, we present the control system of CycleGAN in Figure \ref{Figure11}. CycleGAN has two Generators and two Discriminators, and taking the cycle consistency loss into account, its loss function has three parts as below:

\begin{equation} \label{Eq53}
\begin{aligned}
\mathcal{L}(G_{a},G_{b},D_{a},D_{b}) & = \mathcal{L}_{GAN}(G_{a},D_{b},A,B) + \mathcal{L}_{GAN}(G_{b},D_{a},B,A) + \lambda \mathcal{L}_{cyc}(G_{a},G_{b})
\end{aligned}
\end{equation}

where for the mapping function $G_{a}$: $A \rightarrow B$ and its discriminator $D_{b}$, we express the objective as:

\begin{equation} \label{Eq54}
\begin{aligned}
 \mathcal{L}_{GAN}(G_{a},D_{b},A,B) = \mathbb{E}_{a \thicksim p_{data}(a)} [log D_{b} (b)] + \mathbb{E}_{b \thicksim p_{data}(b)}[log (1+D_{b}(G(a))]
\end{aligned}
\end{equation}

For each image $b$ from domain $B$ , $G_{a}$ and $G_{b}$ should satisfy backward cycle consistency: $b \rightarrow G_{a}(b) \rightarrow G_{a}(G_{b}(b)) \approx y$. Thus, the cycle consistency loss should be:

\begin{equation} \label{Eq55}
\begin{aligned}
 \mathcal{L}_{cyc}(G_{a},G_{b}) = \mathbb{E}_{a \thicksim p_{data}(a)} [|| G_{b}(G_{a}(a))-a ||_{1}] + \mathbb{E}_{b \thicksim p_{data}(b)}[|| G_{a}(G_{b}(b))-b ||_{1}]
\end{aligned}
\end{equation}

CycleGAN used the L1 norm in this loss with an adversarial loss between $G_{b}(G_{a}(a))$ and $a$, and between $G_{a}(G_{b}(b))$ and $b$, but did not observe improved performance. Therefore, we get the system function of CycleGAN as below:

\begin{equation} \label{Eq56}
\theta_{Da}(s) = controller \cdot \theta_{Ga}(s) \cdot E(s)
\end{equation}

\begin{equation} \label{Eq57}
\theta_{Ga}(s) = controller \cdot E(s)
\end{equation} 

\begin{equation} \label{Eq58}
\theta_{Db}(s) = controller \cdot \theta_{Gb}(s) \cdot E(s)
\end{equation}

\begin{equation} \label{Eq59}
\theta_{Gb}(s) = controller \cdot E(s)
\end{equation} 

\begin{equation} \label{Eq60}
\begin{aligned}
E(s) = &\left[ \frac{\theta_{Da}^{\ast}}{s} - \theta_{Da}(s) \right] + \left[ \frac{\theta_{Db}^{\ast}}{s} - \theta_{Db}(s) \right] + \left[ \frac{\theta_{Ga}^{\ast}}{s} - \theta_{Ga}(s)\theta_{Gb}(s) \right] + \left[ \frac{\theta_{Gb}^{\ast}}{s} - \theta_{Gb}(s)\theta_{Ga}(s) \right]\\
\end{aligned}
\end{equation}

We simulated the system response of an advanced GAN -- CycleGAN on seven controllers (optimisers) and summarized the result in Figure \ref{Figure12}. PID and FuzzyPID controllers can generate the excellent stable sinusoidal signals both on $G_{a}$ and $G_{b}$. SGDM controller failed to generate sinusoidal signals, otherwise, SGD and AdaM can generate acceptable sinusoidal signals. For the generated MNIST in Figure \ref{Figure13}, after 100 epochs training, PID can generate $100\%$ correct samples both from $G_{a}$ to $G_{b}$ and from $G_{b}$ to $G_{a}$. Notably, the ability of CycleGAN to produce samples from a single dataset was significantly enhanced when utilizing the FuzzyPID, which yielded flawless samples from the outset. This suggests that FuzzyPID might be the optimal choice for optimizing the learning updates of CycleGAN. The generated samples are depicted in Figure \ref{Figure13}. A manual evaluation of the alignment between samples from and vice versa was also conducted. Preliminary observations indicate that the PID and FuzzyPID optimisers outshine the others when applied to models that utilize a cycle consistency loss, such as CycleGAN.

\begin{figure}[h]
    \centering 
    \begin{subfigure}[b]{0.1\textwidth}
        \caption{$G_{a}$ to $G_{b}$ on the $1_{st}$ epoch.} 
        \centering 
        \includegraphics[scale=0.40]{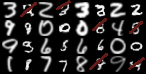} 
        % \label{fig:mean and std of net14} 
    \end{subfigure} 
    \hfill 
    \begin{subfigure}[b]{0.1\textwidth}
        \caption{$G_{a}$ to $G_{b}$ on the $50_{th}$ epoch.} 
        \centering 
        \includegraphics[scale=0.40]{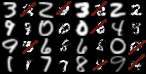} 
        % \label{fig:mean and std of net24} 
    \end{subfigure}
    \hfill
    \begin{subfigure}[b]{0.1\textwidth} 
        \caption{$G_{a}$ to $G_{b}$ on the $100_{th}$ epoch.} 
        \centering 
        \includegraphics[scale=0.40]{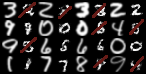} 
        % \label{fig:mean and std of net14} 
    \end{subfigure} 
    \hfill 
    \begin{subfigure}[b]{0.1\textwidth} 
        \caption{$G_{a}$ to $G_{b}$ on the $200_{th}$ epoch.} 
        \centering 
        \includegraphics[scale=0.40]{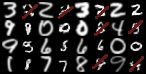} 
        % \label{fig:mean and std of net24} 
    \end{subfigure}
    \hfill 
    \begin{subfigure}[b]{0.1\textwidth} 
        \caption{$G_{b}$ to $G_{a}$ on the $1_{st}$ epoch.} 
        \centering 
        \includegraphics[scale=0.40]{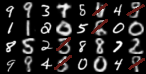} 
        % \label{fig:mean and std of net14} 
    \end{subfigure} 
    \hfill 
    \begin{subfigure}[b]{0.1\textwidth} 
        \caption{$G_{b}$ to $G_{a}$ on the $50_{th}$ epoch.} 
        \centering 
        \includegraphics[scale=0.40]{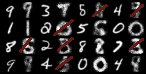} 
        % \label{fig:mean and std of net24} 
    \end{subfigure}
    \hfill
    \begin{subfigure}[b]{0.1\textwidth} 
        \caption{$G_{b}$ to $G_{a}$ on the $100_{th}$ epoch.} 
        \centering 
        \includegraphics[scale=0.40]{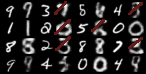} 
        % \label{fig:mean and std of net14} 
    \end{subfigure} 
    \hfill 
    \begin{subfigure}[b]{0.1\textwidth} 
        \caption{$G_{b}$ to $G_{a}$ on the $200_{th}$ epoch.} 
        \centering 
        \includegraphics[scale=0.40]{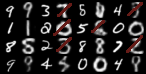} 
        % \label{fig:mean and std of net24} 
    \end{subfigure} 
    \vskip 
    \baselineskip  
    \begin{subfigure}[b]{0.1\textwidth} 
        \centering 
        \includegraphics[scale=0.40]{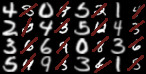} 
        % \caption{\small}
        % \label{fig:mean and std of net14} 
    \end{subfigure} 
    \hfill 
    \begin{subfigure}[b]{0.1\textwidth} 
        \centering 
        \includegraphics[scale=0.40]{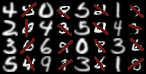} 
        % \caption{\small} 
        % \label{fig:mean and std of net24} 
    \end{subfigure}
    \hfill
    \begin{subfigure}[b]{0.1\textwidth} 
        \centering 
        \includegraphics[scale=0.40]{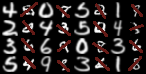} 
        % \caption{\small} 
        % \label{fig:mean and std of net14} 
    \end{subfigure} 
    \hfill 
    \begin{subfigure}[b]{0.1\textwidth} 
        \centering 
        \includegraphics[scale=0.40]{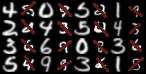} 
        % \caption{\small} 
        % \label{fig:mean and std of net24} 
    \end{subfigure}
    \hfill 
    \begin{subfigure}[b]{0.1\textwidth} 
        \centering 
        \includegraphics[scale=0.40]{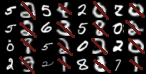} 
        % \caption{\small}
        % \label{fig:mean and std of net14} 
    \end{subfigure} 
    \hfill 
    \begin{subfigure}[b]{0.1\textwidth} 
        \centering 
        \includegraphics[scale=0.40]{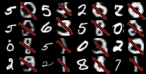}
        % \caption{\small} 
        % \label{fig:mean and std of net24} 
    \end{subfigure}
    \hfill
    \begin{subfigure}[b]{0.1\textwidth} 
        \centering 
        \includegraphics[scale=0.40]{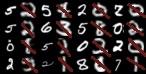} 
        % \caption{\small} 
        % \label{fig:mean and std of net14} 
    \end{subfigure} 
    \hfill 
    \begin{subfigure}[b]{0.1\textwidth} 
        \centering 
        \includegraphics[scale=0.40]{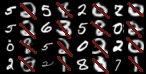} 
        % \caption{\small} 
        % \label{fig:mean and std of net24} 
    \end{subfigure} 
    \vskip 
    \baselineskip  
    \begin{subfigure}[b]{0.1\textwidth} 
        \centering 
        \includegraphics[scale=0.40]{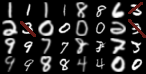} 
        % \caption{\small}
        % \label{fig:mean and std of net14} 
    \end{subfigure} 
    \hfill 
    \begin{subfigure}[b]{0.1\textwidth} 
        \centering 
        \includegraphics[scale=0.40]{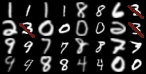} 
        % \caption{\small} 
        % \label{fig:mean and std of net24} 
    \end{subfigure}
    \hfill
    \begin{subfigure}[b]{0.1\textwidth} 
        \centering 
        \includegraphics[scale=0.40]{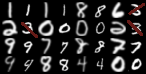} 
        % \caption{\small} 
        % \label{fig:mean and std of net14} 
    \end{subfigure} 
    \hfill 
    \begin{subfigure}[b]{0.1\textwidth} 
        \centering 
        \includegraphics[scale=0.40]{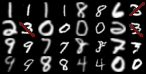} 
        % \caption{\small} 
        % \label{fig:mean and std of net24} 
    \end{subfigure}
    \hfill 
    \begin{subfigure}[b]{0.1\textwidth} 
        \centering 
        \includegraphics[scale=0.40]{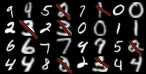} 
        % \caption{\small}
        % \label{fig:mean and std of net14} 
    \end{subfigure} 
    \hfill 
    \begin{subfigure}[b]{0.1\textwidth} 
        \centering 
        \includegraphics[scale=0.40]{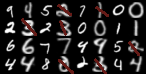} 
        % \caption{\small} 
        % \label{fig:mean and std of net24} 
    \end{subfigure}
    \hfill
    \begin{subfigure}[b]{0.1\textwidth} 
        \centering 
        \includegraphics[scale=0.40]{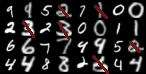} 
        % \caption{\small} 
        % \label{fig:mean and std of net14} 
    \end{subfigure} 
    \hfill 
    \begin{subfigure}[b]{0.1\textwidth} 
        \centering 
        \includegraphics[scale=0.40]{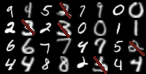} 
        % \caption{\small} 
        % \label{fig:mean and std of net24} 
    \end{subfigure}
    \vskip 
    \baselineskip  
    \begin{subfigure}[b]{0.1\textwidth} 
        \centering 
        \includegraphics[scale=0.40]{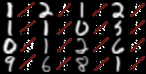} 
        % \caption{\small}
        % \label{fig:mean and std of net14} 
    \end{subfigure} 
    \hfill 
    \begin{subfigure}[b]{0.1\textwidth} 
        \centering 
        \includegraphics[scale=0.40]{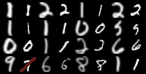} 
        % \caption{\small} 
        % \label{fig:mean and std of net24} 
    \end{subfigure}
    \hfill
    \begin{subfigure}[b]{0.1\textwidth} 
        \centering 
        \includegraphics[scale=0.40]{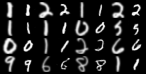} 
        % \caption{\small} 
        % \label{fig:mean and std of net14} 
    \end{subfigure} 
    \hfill 
    \begin{subfigure}[b]{0.1\textwidth} 
        \centering 
        \includegraphics[scale=0.40]{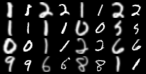} 
        % \caption{\small} 
        % \label{fig:mean and std of net24} 
    \end{subfigure}
    \hfill 
    \begin{subfigure}[b]{0.1\textwidth} 
        \centering 
        \includegraphics[scale=0.40]{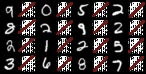} 
        % \caption{\small}
        % \label{fig:mean and std of net14} 
    \end{subfigure} 
    \hfill 
    \begin{subfigure}[b]{0.1\textwidth} 
        \centering 
        \includegraphics[scale=0.40]{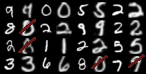} 
        % \caption{\small} 
        % \label{fig:mean and std of net24} 
    \end{subfigure}
    \hfill
    \begin{subfigure}[b]{0.1\textwidth} 
        \centering 
        \includegraphics[scale=0.40]{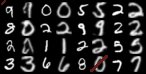} 
        % \caption{\small} 
        % \label{fig:mean and std of net14} 
    \end{subfigure} 
    \hfill 
    \begin{subfigure}[b]{0.1\textwidth} 
        \centering 
        \includegraphics[scale=0.40]{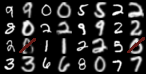} 
        % \caption{\small} 
        % \label{fig:mean and std of net24} 
    \end{subfigure}
            \vskip 
    \baselineskip  
    \begin{subfigure}[b]{0.1\textwidth} 
        \centering 
        \includegraphics[scale=0.40]{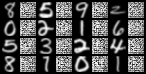} 
        % \caption{\small}
        % \label{fig:mean and std of net14} 
    \end{subfigure} 
    \hfill 
    \begin{subfigure}[b]{0.1\textwidth} 
        \centering 
        \includegraphics[scale=0.40]{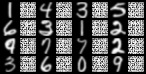} 
        % \caption{\small} 
        % \label{fig:mean and std of net24} 
    \end{subfigure}
    \hfill
    \begin{subfigure}[b]{0.1\textwidth} 
        \centering 
        \includegraphics[scale=0.40]{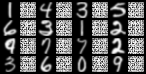} 
        % \caption{\small} 
        % \label{fig:mean and std of net14} 
    \end{subfigure} 
    \hfill 
    \begin{subfigure}[b]{0.1\textwidth} 
        \centering 
        \includegraphics[scale=0.40]{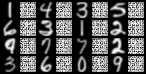} 
        % \caption{\small} 
        % \label{fig:mean and std of net24} 
    \end{subfigure}
    \hfill 
    \begin{subfigure}[b]{0.1\textwidth} 
        \centering 
        \includegraphics[scale=0.40]{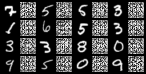} 
        % \caption{\small}
        % \label{fig:mean and std of net14} 
    \end{subfigure} 
    \hfill 
    \begin{subfigure}[b]{0.1\textwidth} 
        \centering 
        \includegraphics[scale=0.40]{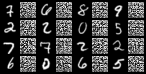} 
        % \caption{\small} 
        % \label{fig:mean and std of net24} 
    \end{subfigure}
    \hfill
    \begin{subfigure}[b]{0.1\textwidth} 
        \centering 
        \includegraphics[scale=0.40]{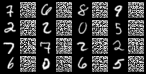} 
        % \caption{\small} 
        % \label{fig:mean and std of net14} 
    \end{subfigure} 
    \hfill 
    \begin{subfigure}[b]{0.1\textwidth} 
        \centering 
        \includegraphics[scale=0.40]{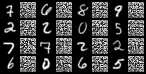} 
        % \caption{\small} 
        % \label{fig:mean and std of net24} 
    \end{subfigure}
    \vskip
    \baselineskip  
    \begin{subfigure}[b]{0.1\textwidth} 
        \centering 
        \includegraphics[scale=0.40]{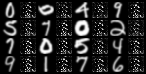} 
        % \caption{\small}
        % \label{fig:mean and std of net14} 
    \end{subfigure} 
    \hfill 
    \begin{subfigure}[b]{0.1\textwidth} 
        \centering 
        \includegraphics[scale=0.40]{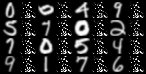} 
        % \caption{\small} 
        % \label{fig:mean and std of net24} 
    \end{subfigure}
    \hfill
    \begin{subfigure}[b]{0.1\textwidth} 
        \centering 
        \includegraphics[scale=0.40]{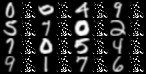} 
        % \caption{\small} 
        % \label{fig:mean and std of net14} 
    \end{subfigure} 
    \hfill 
    \begin{subfigure}[b]{0.1\textwidth} 
        \centering 
        \includegraphics[scale=0.40]{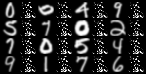} 
        % \caption{\small} 
        % \label{fig:mean and std of net24} 
    \end{subfigure}
    \hfill 
    \begin{subfigure}[b]{0.1\textwidth} 
        \centering 
        \includegraphics[scale=0.40]{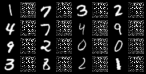} 
        % \caption{\small}
        % \label{fig:mean and std of net14} 
    \end{subfigure} 
    \hfill 
    \begin{subfigure}[b]{0.1\textwidth} 
        \centering 
        \includegraphics[scale=0.40]{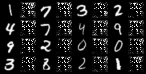} 
        % \caption{\small} 
        % \label{fig:mean and std of net24} 
    \end{subfigure}
    \hfill
    \begin{subfigure}[b]{0.1\textwidth} 
        \centering 
        \includegraphics[scale=0.40]{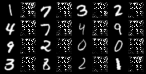} 
        % \caption{\small} 
        % \label{fig:mean and std of net14} 
    \end{subfigure} 
    \hfill 
    \begin{subfigure}[b]{0.1\textwidth} 
        \centering 
        \includegraphics[scale=0.40]{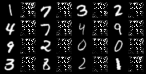} 
        % \caption{\small} 
        % \label{fig:mean and std of net24} 
    \end{subfigure}
    \vskip 
    \baselineskip  
    \begin{subfigure}[b]{0.1\textwidth} 
        \centering 
        \includegraphics[scale=0.40]{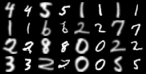} 
        % \caption{\small}
        % \label{fig:mean and std of net14} 
    \end{subfigure} 
    \hfill 
    \begin{subfigure}[b]{0.1\textwidth} 
        \centering 
        \includegraphics[scale=0.40]{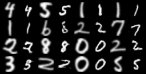} 
        % \caption{\small} 
        % \label{fig:mean and std of net24} 
    \end{subfigure}
    \hfill
    \begin{subfigure}[b]{0.1\textwidth} 
        \centering 
        \includegraphics[scale=0.40]{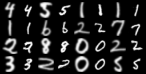} 
        % \caption{\small} 
        % \label{fig:mean and std of net14} 
    \end{subfigure} 
    \hfill 
    \begin{subfigure}[b]{0.1\textwidth} 
        \centering 
        \includegraphics[scale=0.40]{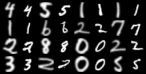} 
        % \caption{\small} 
        % \label{fig:mean and std of net24} 
    \end{subfigure}
    \hfill 
    \begin{subfigure}[b]{0.1\textwidth} 
        \centering 
        \includegraphics[scale=0.40]{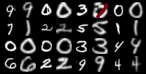} 
        % \caption{\small}
        % \label{fig:mean and std of net14} 
    \end{subfigure} 
    \hfill 
    \begin{subfigure}[b]{0.1\textwidth} 
        \centering 
        \includegraphics[scale=0.40]{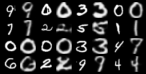} 
        % \caption{\small} 
        % \label{fig:mean and std of net24} 
    \end{subfigure}
    \hfill
    \begin{subfigure}[b]{0.1\textwidth} 
        \centering 
        \includegraphics[scale=0.40]{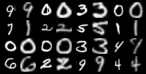} 
        % \caption{\small} 
        % \label{fig:mean and std of net14} 
    \end{subfigure} 
    \hfill 
    \begin{subfigure}[b]{0.1\textwidth} 
        \centering 
        \includegraphics[scale=0.40]{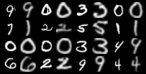} 
        % \caption{\small} 
        % \label{fig:mean and std of net24} 
    \end{subfigure}
    \caption{The generated samples from CycleGAN on corresponding optimisers (From top to bottom is SGD, SGDM, AdaM, PID, LPF-SGD, HPF-SGD and FuzzyPID).} 
    \label{Figure13} 
\end{figure}

\section{Hyperparameters}

\begin{table}[h]
\caption{Hyper-parameters for the image classification task on MNIST.}
\label{Table3}
\scriptsize
\centering
\begin{tabular}{cccc}
\toprule
\textbf{Hyper-parameter}    & \textbf{\begin{tabular}[c]{@{}c@{}}Backward System\end{tabular}} & \textbf{\begin{tabular}[c]{@{}c@{}}Forward  System\end{tabular}} & \textbf{\begin{tabular}[c]{@{}c@{}}Backward-Forward System\end{tabular}} \\ \hline
Data augmentation                 & Auto  & Auto  & Auto  \\
Input resolution                  & [28,28,1]  & [28,28,1]  & [28,28,1]  \\
Epochs                            & 40   & 200  & 200  \\
Batch size                        & 100  & 100  & 100  \\
Hidden dropout                    & 0 & 0  & 0\\
Random erasing prob               & 0  & 0  & 0 \\
EMA decay                         & 0  & 0  & 0  \\
Cutmix $\alpha$                   & 0  & 0  & 0  \\
Mixup $\alpha$                    & 0  & 0  & 0  \\
Cutmix-Mixup                      & 0  & 0  & 0 \\
Label smoothing                   & 0.1  & 0.1  & 0.1 \\
Peak learning rate                & 2e-3  & 2e-3  & 2e-4  \\
Steps per block                & /  & 60  & /  \\
Positive samples portion $\lambda$                   & /  & [0.3, 0.5, 0.7]  & /  \\
$Threshold$                & /  & [0.1, 1.0, 10.0]  & /  \\
optimiser                         & \multicolumn{3}{c}{\{SGD, SGDM, AdaM, PID, LPF-SGD, HPF-SGD, FuzzyPID\}}  \\
\bottomrule
\end{tabular}
\end{table}

In this study, we conducted three primary experiments, as detailed in Table \ref{Table3}. We roughly separated them to experiments on Backward System, Forward  System, Backward-Forward System. Experiments on Backward System and Backward-Forward System do not have Steps per block, $Threshold$ and Positive samples portion. However, to make a fair comparison, all experiment should use seven optimisers on the same hyperparameters. These experiments were categorised based on the Backward System, Forward System, and Backward-Forward System. Notably, the Backward System and Backward-Forward System do not utilise the "Steps per block", "$Threshold$", and "Positive samples portion" hyperparameters. Nonetheless, for a rigorous comparison, all experiments employed the same seven optimisers with consistent hyperparameters.

\begin{table}[h]
\caption{Hyper-parameters for the image classification task on CIFAR10, CIFAR100 and TinyImageNet.}
\label{Table4}
\scriptsize
\centering
\begin{tabular}{p{0.25\linewidth} p{0.05\linewidth} p{0.06\linewidth} p{0.06\linewidth} p{0.06\linewidth} 
p{0.08\linewidth} p{0.08\linewidth} p{0.08\linewidth} }
\toprule
\textbf{Hyper-parameter}    & VGG19 & ResNet18 & ResNet50 & ResNet101 & DenseNet121 & MobileNetV2 & EffecientNet \\ \hline
Data augmentation                 & Auto  & Auto  & Auto  & Auto  & Auto  & Auto  & Auto \\
Input resolution (CIFAR10,100)                  & [32,32,3]  & [32,32,3]  & [32,32,3]  & [32,32,3]  & [32,32,3]  & [32,32,3]  & [32,32,3] \\
Input resolution (TinyImageNet)                  & [64,64,3]  & [64,64,3]  & [64,64,3]  & [64,64,3]  & [64,64,3]  & [64,64,3]  & [64,64,3] \\
Epochs                            & 200   & 200  & 200 & 200   & 200  & 200 & 200  \\
Batch size                        & 100  & 100  & 100 & 100  & 100  & 100 & 100   \\
Hidden dropout                    & 0 & 0 & 0  & 0  & 0 & 0 & 0\\
Random erasing prob               & 0  & 0  & 0  & 0 & 0 & 0  & 0\\
EMA decay                         & 0  & 0  & 0   & 0 & 0 & 0  & 0\\
Cutmix $\alpha$                   & 0  & 0  & 0   & 0 & 0 & 0  & 0\\
Mixup $\alpha$                    & 0  & 0  & 0   & 0 & 0 & 0  & 0\\
Cutmix-Mixup                      & 0  & 0  & 0  & 0 & 0 & 0  & 0\\
Label smoothing                   & 0.1  & 0.1  & 0.1 & 0.1  & 0.1  & 0.1  & 0.1\\
Peak learning rate                & 2e-2  & 2e-2  & 2e-2  & 2e-2  & 2e-2  & 2e-2  & 2e-2\\
optimiser                         & \multicolumn{7}{c}{\{SGD, SGDM, AdaM, PID, LPF-SGD, HPF-SGD, FuzzyPID\}}  \\
\bottomrule
\end{tabular}
\end{table}

As shown in Table \ref{Table4}, we employed three datasets: CIFAR10, CIFAR100 and TinyImageNet. Additionally, one vision model -- VGG19 which lacks the residual connection, and six residual connections used vision models are illustrated in our experiments. We specifically chose these seven vision models to investigate whether a more complex system, indicating a learning system order of two or higher, can be ascertained.

\begin{table}[h]
\caption{Coefficients of LPF-SGD and HPF-SGD using second-order IIR structure.}
\label{Table5}
\scriptsize
\centering
\begin{tabular}{cccccccc}
\toprule
\multirow{2}{*}{\textbf{Filter Type}}   & Gain  & \multicolumn{3}{c}{Numerator} & \multicolumn{3}{c}{Denominator} \\ \cline{2-8}
 & $G$ & $x_{0}$ & $x_{1}$  & $x_{2}$  & $y_{0}$  & $y_{1}$  & $y_{2}$\\ \hline
Low Pass Filter             & 0.49968  & 1 & -0.99937 & 0.00063 & 1.0  & 0  & -1.0   \\ 
High Pass Filter             & 0.49968  & 1 & 0.99937 & 0.00063 & 1.0  & 0  & -1.0   \\
\bottomrule
\end{tabular}
\end{table}

Meanwhile, we designed two filter processed SGD optimisers by using a second-order IIR structure. The coefficient of the convolution process in Equation \ref{Eq31} is listed in Table \ref{Table5}. Owing to the frequency cutoff around the midpoint (given the uncertainty in determining the sampling rate and the desired frequency band), this second-order IIR filter encompasses seven coefficients. It's noteworthy that the \textbf{'filterDesigner'} toolbox in MATLAB can be utilized to design such filters.

\section{CIFAR10, CIFAR100 and TinyImageNet}

This section presents the accuracy rate using seven vision models (e.g., VGG19, ResNet18, ResNet50, ResNet101, DenseNet121, MobileNetV2 and EfficientNet) across seven optimisers (SGD, SGDM, AdaM , PID, LPF-SGD, HPF-SGD, and FuzzyPID). These results are detailed in Table \ref{Table6}, and the associated training and testing curves are depicted in Figure \ref{Figure14}. VGG19 is a straight-forward connected vision model without residual blocks, and as demonstrated by the system response in Figure \ref{Figure9}, no matter the assumed system order of VGG19 is one or two, compared to SGD on ResNet50, SGD on VGG19 only can achieve the half accuracy rate on CIFAR100 and TinyImageNet. Introducing a low pass filter to SGD results in a considerably slow learning curve ascent. Conversely, incorporating a high pass filter facilitates the learning process. We infer that the update of weights needs the high frequency component of gradient sequences to rapidly adapt to the optimal. Consistently, because of the adaptive part $M$ in Equation \ref{Eq7}, AdaM aims to follow the change of gradients with a faster speed. Nonetheless, relying solely on a single parameter, $\beta_{2}$ , for updates does not effectively mitigate the overshoot issue. Interestingly, the FuzzyPID optimiser exhibits a smoother learning trajectory compared to the PID. The design intention behind FuzzyPID was to supplement the PID optimiser, aiding in the adjustment of its overshoot issue. However, in practice, while FuzzyPID may not consistently outperform PID, it exhibits superior performance when deployed on CycleGAN.

\begin{table}[h]
\caption{The results of CNN with different optimisers on CIFAR10 and CIFAR100. Using the 10-fold cross-validation, the average and standard variance results are shown below. }
\centering
\label{Table6}
\scriptsize
\begin{tabular}{p{0.14\linewidth} p{0.08\linewidth} p{0.08\linewidth} p{0.08\linewidth} p{0.08\linewidth} p{0.08\linewidth} p{0.08\linewidth} p{0.08\linewidth} p{0.09\linewidth}}
\toprule
\textbf{optimiser}  &SGD  &SGDM  &Adam  &PID  &LPF-SGD  &HPF-SGD  &FuzzyPID  \\
% \textbf{Membership}  &/  &/  &/  &/  &/  &/ &Gaussian \\
% \textbf{Universe}  &/  &/  &/  &/  &/  &/  &[-0.02 0.02]
 % \\ 
\hline
\multicolumn{8}{c}{CIFAR10} \\ \hline
\textbf{$VGG19$} & $90.89_{\pm0.03}$ &$93.13_{\pm0.13}$  &$78.07_{\pm0.82}$  &$93.60_{\pm0.06}$  &$13.75_{\pm0.69}$  &$92.52_{\pm0.09}$  &$93.45_{\pm0.09}$  \\
\textbf{$ResNet18$} & $91.65_{\pm0.15}$ &$94.67_{\pm0.17}$  &$83.54_{\pm1.32}$  &$95.42_{\pm0.05}$  &$11.07_{\pm2.09}$  &$93.05_{\pm0.12}$  &$94.98_{\pm0.19}$  \\
\textbf{$ResNet50$}  &$91.06_{\pm0.02}$ &$94.70_{\pm0.12}$  &$81.39_{\pm0.44}$  &$95.21_{\pm0.19}$  &$11.49_{\pm0.06}$   &$92.84_{\pm0.09}$  &$94.51_{\pm0.35}$  \\
\textbf{$ResNet101$}  &$90.87_{\pm0.29}$ &$94.70_{\pm0.08}$  &$82.78_{\pm0.05}$  &$95.39_{\pm0.14}$  &$10.49_{\pm0.14}$  &$92.42_{\pm0.11}$  &$93.95_{\pm0.20}$  \\
\textbf{$DenseNet121$}  &$91.37_{\pm0.15}$ &$95.20_{\pm0.28}$  &$84.62_{\pm0.27}$  &$95.71_{\pm0.04}$  &$11.90_{\pm0.69}$ &$93.23_{\pm0.07}$ &$94.53_{\pm0.13}$ \\
\textbf{$MobileNetV2$}  &$87.97_{\pm0.07}$ &$93.78_{\pm0.01}$  &$83.02_{\pm0.24}$  &$94.12_{\pm0.25}$  &$10.02_{\pm0.75}$  &$90.74_{\pm0.12}$  &$93.46_{\pm0.03}$  \\
\textbf{$EffecientNet$}  &$83.36_{\pm0.48}$ &$91.71_{\pm0.26}$  &$83.49_{\pm0.13}$  &$92.14_{\pm1.10}$  &$10.69_{\pm0.61}$ &$87.66_{\pm0.13}$  &$91.16_{\pm0.15}$ \\
\hline
\multicolumn{8}{c}{CIFAR100} \\ \hline
\textbf{$VGG19$} & $66.42_{\pm0.28}$ &$71.97_{\pm0.21}$  &$18.28_{\pm2.12}$  &$73.25_{\pm0.05}$  &$1.35_{\pm0.05}$  &$69.65_{\pm0.21}$  &$73.21_{\pm0.29}$  \\
\textbf{$ResNet18$} & $70.05_{\pm0.03}$ &$76.03_{\pm0.02}$  &$49.90_{\pm0.02}$  &$77.84_{\pm0.01}$  &$1.12_{\pm0.02}$  &$72.54_{\pm0.17}$  &$76.78_{\pm0.08}$  \\
\textbf{$ResNet50$}  &$66.52_{\pm0.42}$ &$77.25_{\pm0.16}$  &$49.64_{\pm0.80}$  &$78.37_{\pm0.04}$  &$1.28_{\pm0.10}$  &$71.23_{\pm0.10}$  &$75.62_{\pm0.16}$  \\
\textbf{$ResNet101$}  &$64.69_{\pm0.11}$ &$76.36_{\pm0.38}$  &$50.97_{\pm0.91}$  &$79.25_{\pm0.26}$  &$1.09_{\pm0.04}$  &$70.61_{\pm0.14}$  &$72.78_{\pm0.31}$  \\
\textbf{$DenseNet121$}  &$68.45_{\pm0.39}$ &$77.78_{\pm0.31}$  &$55.87_{\pm0.82}$  &$80.06_{\pm0.12}$  &$1.12_{\pm0.16}$  &$73.72_{\pm0.21}$  &$75.71_{\pm0.25}$  \\
\textbf{$MobileNetV2$}  &$62.52_{\pm0.37}$ &$73.96_{\pm0.26}$  &$42.74_{\pm2.80}$  &$74.81_{\pm0.11}$  &$0.98_{\pm0.02}$  &$67.83_{\pm0.14}$  &$73.31_{\pm0.26}$  \\
\textbf{$EffecientNet$}  &$50.35_{\pm0.77}$ &$66.87_{\pm0.38}$  &$34.06_{\pm7.41}$  &$71.32_{\pm0.29}$  &$1.01_{\pm0.04}$  &$55.98_{\pm0.21}$  &$62.92_{\pm0.35}$  \\
\hline
\multicolumn{8}{c}{TinyImageNet} \\ \hline
\textbf{$VGG19$} & $44.87_{\pm0.01}$ &$51.22_{\pm0.20}$  &$0.57_{\pm0.07}$  &$0.50_{\pm0.00}$  &$0.00_{\pm0.00}$  &$46.66_{\pm0.07}$  &$0.50_{\pm0.00}$\\
\textbf{$ResNet18$} & $50.27_{\pm0.25}$ &$58.79_{\pm0.15}$  &$34.33_{\pm1.23}$  &$63.71_{\pm0.71}$  &$0.00_{\pm0.09}$  &$54.46_{\pm0.09}$  &$61.39_{\pm0.13}$  \\
\textbf{$ResNet50$}  &$42.20_{\pm0.08}$ &$63.54_{\pm0.20}$  &$35.01_{\pm1.18}$  &$67.51_{\pm0.40}$  &$0.00_{\pm0.00}$  &$48.38_{\pm0.10}$  &$60.82_{\pm0.50}$  \\
\textbf{$ResNet101$}  &$40.98_{\pm0.31}$ &$64.13_{\pm0.19}$  &$35.64_{\pm0.71}$  &$69.54_{\pm0.56}$  &$0.40_{\pm0.00}$  &$48.22_{\pm0.06}$  &$58.77_{\pm0.21}$  \\
\textbf{$DenseNet121$}  &$43.23_{\pm0.15}$ &$63.37_{\pm0.07}$  &$38.84_{\pm0.62}$  &$68.29_{\pm0.15}$  &$0.26_{\pm0.02}$  &$49.34_{\pm0.22}$  &$55.66_{\pm0.12}$  \\
\textbf{$MobileNetV2$}  &$46.12_{\pm0.11}$ &$61.08_{\pm0.61}$  &$28.56_{\pm0.28}$  &$59.98_{\pm0.04}$  &$0.00_{\pm0.00}$  &$50.48_{\pm0.12}$  &$58.73_{\pm0.13}$  \\
\textbf{$EffecientNet$}  &$54.12_{\pm0.22}$ &$62.28_{\pm0.04}$  &$23.11_{\pm0.03}$  &$62.66_{\pm0.12}$  &$0.30_{\pm0.02}$  &$57.58_{\pm0.04}$  &$60.43_{\pm0.39}$  \\
\bottomrule
\end{tabular}
\end{table}

\begin{figure}[h]
    \centering 
    \begin{subfigure}[b]{0.21\textwidth}
        \caption{Training Accuracy on CIFAR10.} 
        \centering 
        \includegraphics[scale=0.17]{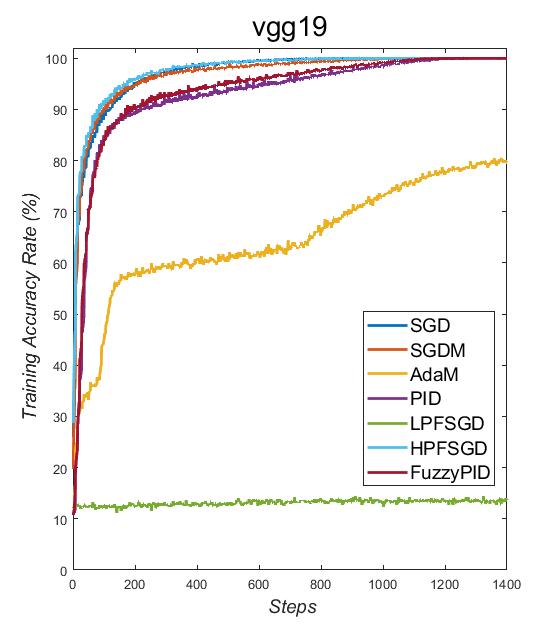} 
        % \label{fig:mean and std of net14} 
    \end{subfigure} 
    \hfill 
    \begin{subfigure}[b]{0.21\textwidth}
        \caption{Testing Accuracy on CIFAR10.} 
        \centering 
        \includegraphics[scale=0.17]{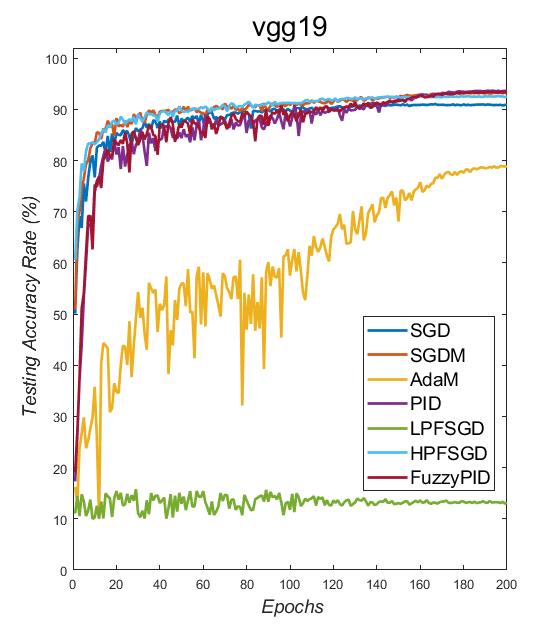} 
        % \label{fig:mean and std of net14} 
    \end{subfigure} 
    \hfill 
    \begin{subfigure}[b]{0.21\textwidth}
        \caption{Training Accuracy on CIFAR100.} 
        \centering 
        \includegraphics[scale=0.17]{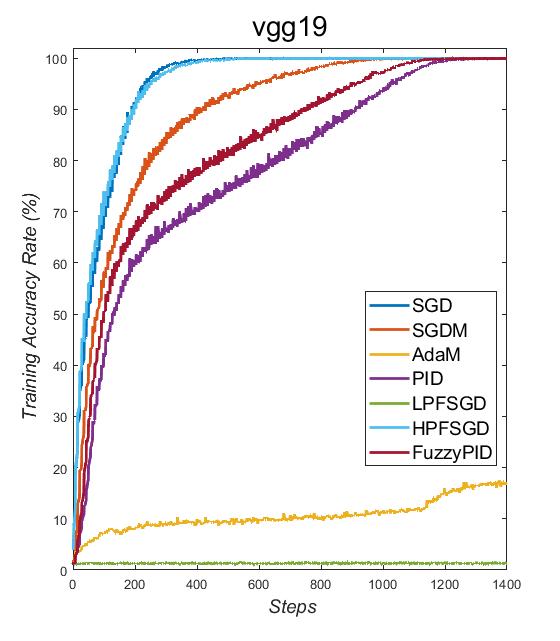} 
        % \label{fig:mean and std of net14} 
    \end{subfigure} 
    \hfill 
    \begin{subfigure}[b]{0.21\textwidth}
        \caption{Testing Accuracy on CIFAR100.} 
        \centering 
        \includegraphics[scale=0.17]{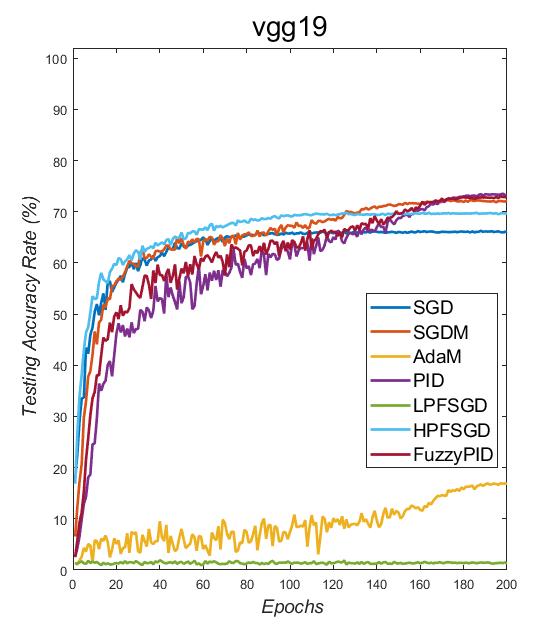} 
        % \label{fig:mean and std of net14} 
    \end{subfigure} 
    % \vskip %\baselineskip 
    % \begin{subfigure}[b]{0.21\textwidth}
    %     \caption{$G_{a}$ to $G_{b}$ on 1 epochs} 
    %     \centering 
    %     \includegraphics[scale=0.17]{VAE_Images/Train_ResNet18} 
    %     % \label{fig:mean and std of net14} 
    % \end{subfigure} 
    % \hfill 
    % \begin{subfigure}[b]{0.21\textwidth}
    %     \caption{$G_{a}$ to $G_{b}$ on 1 epochs} 
    %     \centering 
    %     \includegraphics[scale=0.17]{VAE_Images/Test_ResNet18} 
    %     % \label{fig:mean and std of net14} 
    % \end{subfigure} 
    % \hfill 
    % \begin{subfigure}[b]{0.21\textwidth}
    %     \caption{$G_{a}$ to $G_{b}$ on 1 epochs} 
    %     \centering 
    %     \includegraphics[scale=0.17]{VAE_Images/CIFAR100_Train_ResNet18} 
    %     % \label{fig:mean and std of net14} 
    % \end{subfigure} 
    % \hfill 
    % \begin{subfigure}[b]{0.21\textwidth}
    %     \caption{$G_{a}$ to $G_{b}$ on 1 epochs} 
    %     \centering 
    %     \includegraphics[scale=0.17]{VAE_Images/CIFAR100_Test_ResNet18} 
    %     % \label{fig:mean and std of net14} 
    % \end{subfigure} 
    \vskip 
    \baselineskip 
    \begin{subfigure}[b]{0.21\textwidth}
        % \caption{$G_{a}$ to $G_{b}$ on 1 epochs} 
        \centering 
        \includegraphics[scale=0.17]{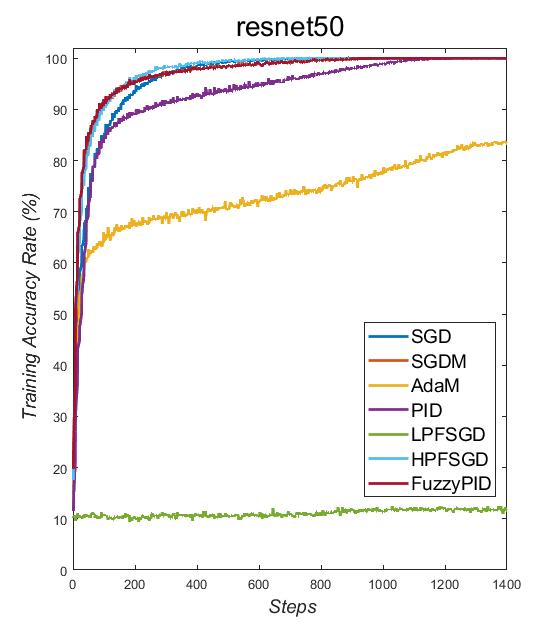} 
        % \label{fig:mean and std of net14} 
    \end{subfigure} 
    \hfill 
    \begin{subfigure}[b]{0.21\textwidth}
        % \caption{$G_{a}$ to $G_{b}$ on 1 epochs} 
        \centering 
        \includegraphics[scale=0.17]{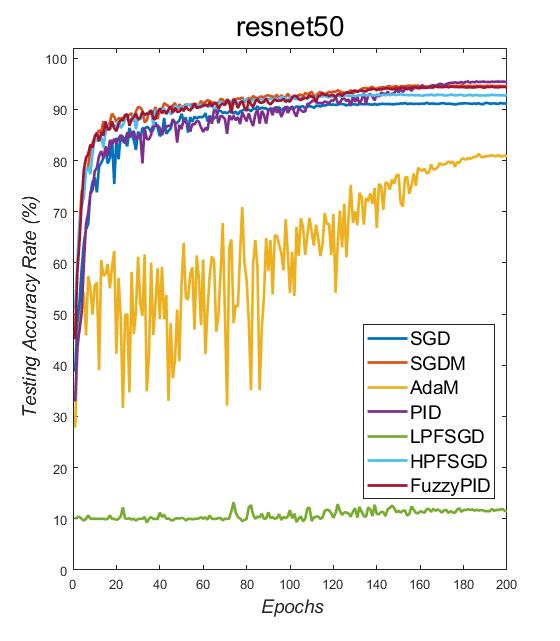} 
        % \label{fig:mean and std of net14} 
    \end{subfigure} 
    \hfill 
    \begin{subfigure}[b]{0.21\textwidth}
        % \caption{$G_{a}$ to $G_{b}$ on 1 epochs} 
        \centering 
        \includegraphics[scale=0.17]{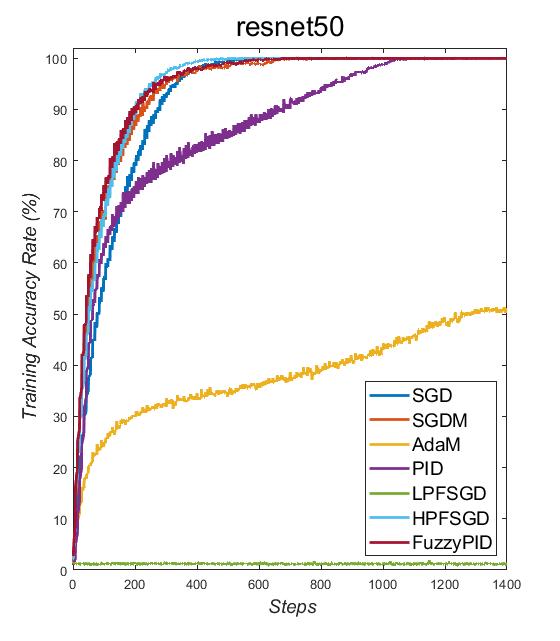} 
        % \label{fig:mean and std of net14} 
    \end{subfigure} 
    \hfill 
    \begin{subfigure}[b]{0.21\textwidth}
        % \caption{$G_{a}$ to $G_{b}$ on 1 epochs} 
        \centering 
        \includegraphics[scale=0.17]{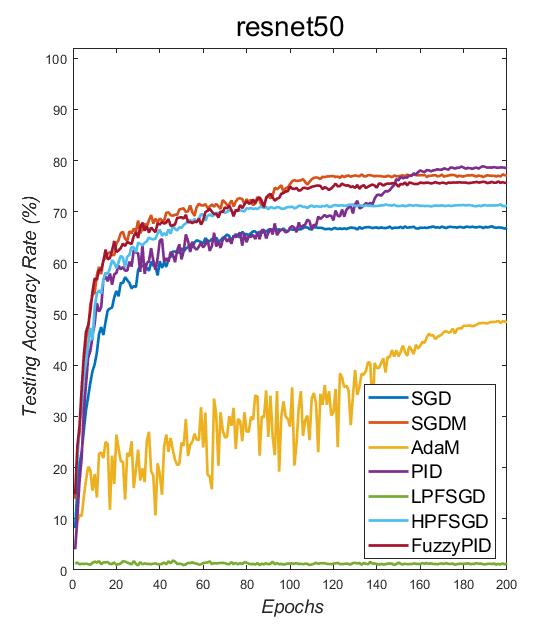} 
        % \label{fig:mean and std of net14} 
    \end{subfigure} 
    % \vskip %\baselineskip 
    % \begin{subfigure}[b]{0.21\textwidth}
    %     \caption{$G_{a}$ to $G_{b}$ on 1 epochs} 
    %     \centering 
    %     \includegraphics[scale=0.17]{VAE_Images/Train_ResNet101} 
    %     % \label{fig:mean and std of net14} 
    % \end{subfigure} 
    % \hfill 
    % \begin{subfigure}[b]{0.21\textwidth}
    %     \caption{$G_{a}$ to $G_{b}$ on 1 epochs} 
    %     \centering 
    %     \includegraphics[scale=0.17]{VAE_Images/Test_ResNet101} 
    %     % \label{fig:mean and std of net14} 
    % \end{subfigure} 
    % \hfill 
    % \begin{subfigure}[b]{0.21\textwidth}
    %     \caption{$G_{a}$ to $G_{b}$ on 1 epochs} 
    %     \centering 
    %     \includegraphics[scale=0.17]{VAE_Images/CIFAR100_Train_ResNet101} 
    %     % \label{fig:mean and std of net14} 
    % \end{subfigure} 
    % \hfill 
    % \begin{subfigure}[b]{0.21\textwidth}
    %     \caption{$G_{a}$ to $G_{b}$ on 1 epochs} 
    %     \centering 
    %     \includegraphics[scale=0.17]{VAE_Images/CIFAR100_Test_ResNet101} 
    %     % \label{fig:mean and std of net14} 
    % \end{subfigure} 
    \vskip 
    \baselineskip 
    \begin{subfigure}[b]{0.21\textwidth}
        % \caption{$G_{a}$ to $G_{b}$ on 1 epochs} 
        \centering 
        \includegraphics[scale=0.17]{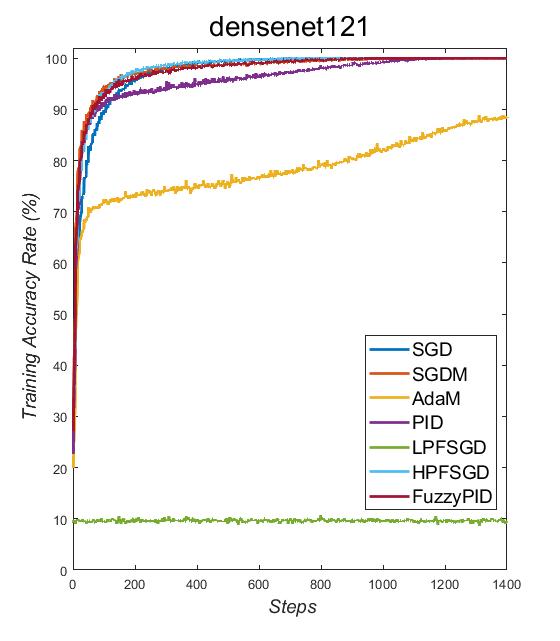} 
        % \label{fig:mean and std of net14} 
    \end{subfigure} 
    \hfill 
    \begin{subfigure}[b]{0.21\textwidth}
        % \caption{$G_{a}$ to $G_{b}$ on 1 epochs} 
        \centering 
        \includegraphics[scale=0.17]{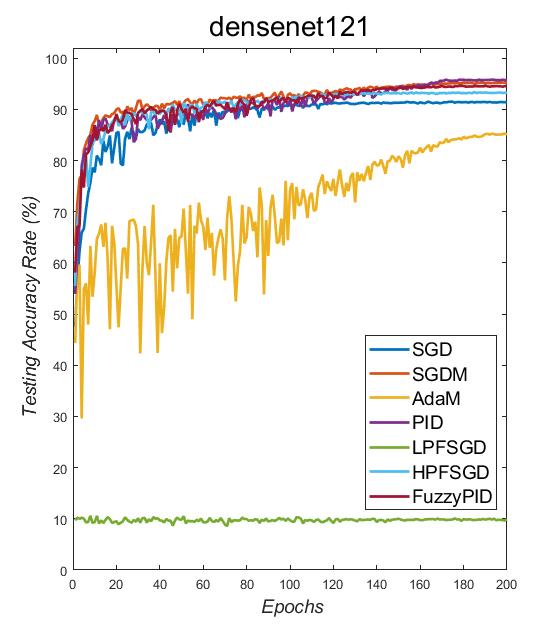} 
        % \label{fig:mean and std of net14} 
    \end{subfigure} 
    \hfill 
    \begin{subfigure}[b]{0.21\textwidth}
        % \caption{$G_{a}$ to $G_{b}$ on 1 epochs} 
        \centering 
        \includegraphics[scale=0.17]{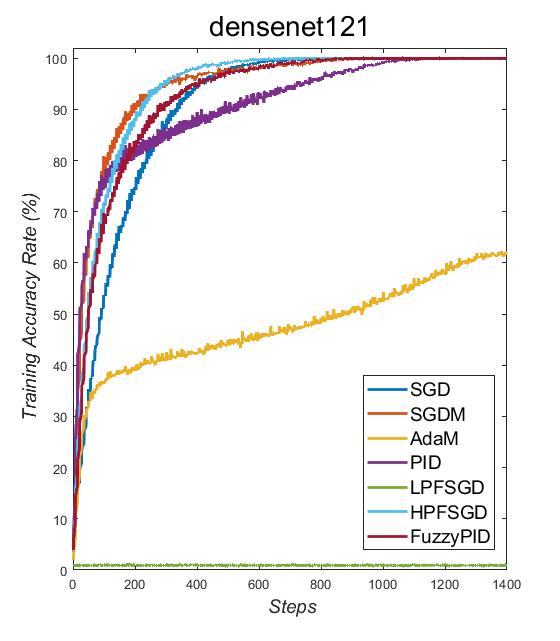} 
        % \label{fig:mean and std of net14} 
    \end{subfigure} 
    \hfill 
    \begin{subfigure}[b]{0.21\textwidth}
        % \caption{$G_{a}$ to $G_{b}$ on 1 epochs} 
        \centering 
        \includegraphics[scale=0.17]{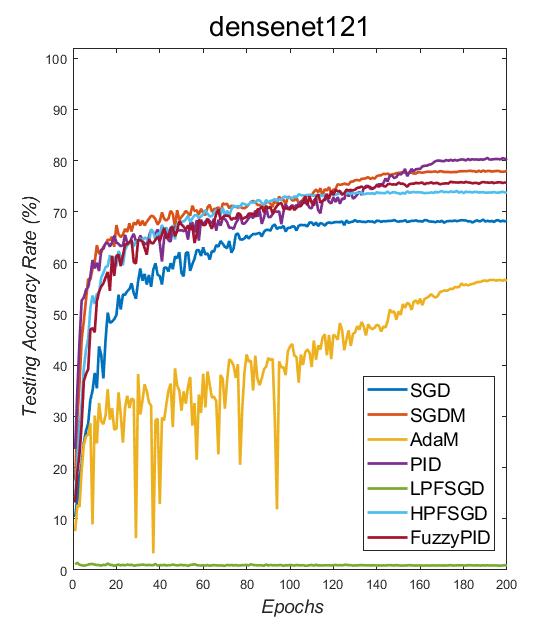} 
        % \label{fig:mean and std of net14} 
    \end{subfigure} 
    \vskip 
    \baselineskip 
    \begin{subfigure}[b]{0.21\textwidth}
        % \caption{$G_{a}$ to $G_{b}$ on 1 epochs} 
        \centering 
        \includegraphics[scale=0.17]{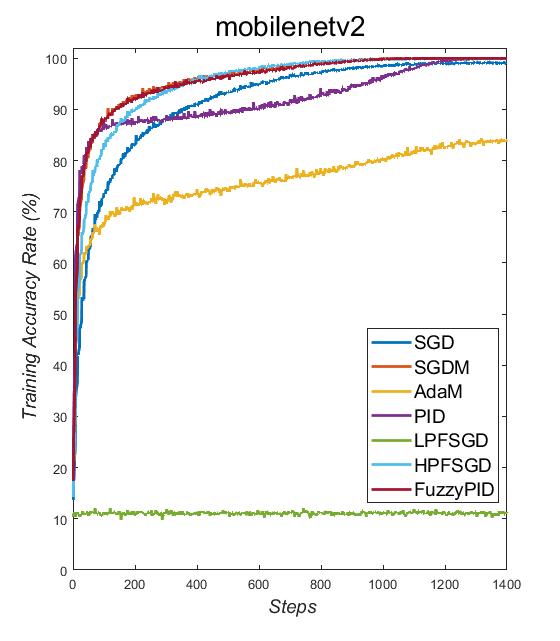} 
        % \label{fig:mean and std of net14} 
    \end{subfigure} 
    \hfill 
    \begin{subfigure}[b]{0.21\textwidth}
        % \caption{$G_{a}$ to $G_{b}$ on 1 epochs} 
        \centering 
        \includegraphics[scale=0.17]{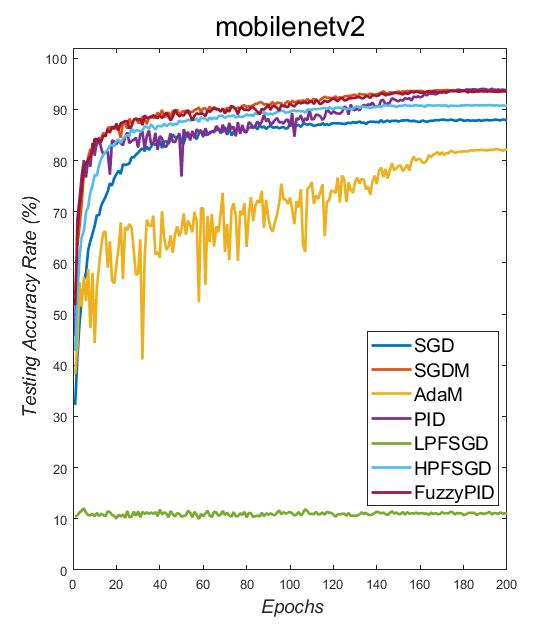} 
        % \label{fig:mean and std of net14} 
    \end{subfigure} 
    \hfill 
    \begin{subfigure}[b]{0.21\textwidth}
        % \caption{$G_{a}$ to $G_{b}$ on 1 epochs} 
        \centering 
        \includegraphics[scale=0.17]{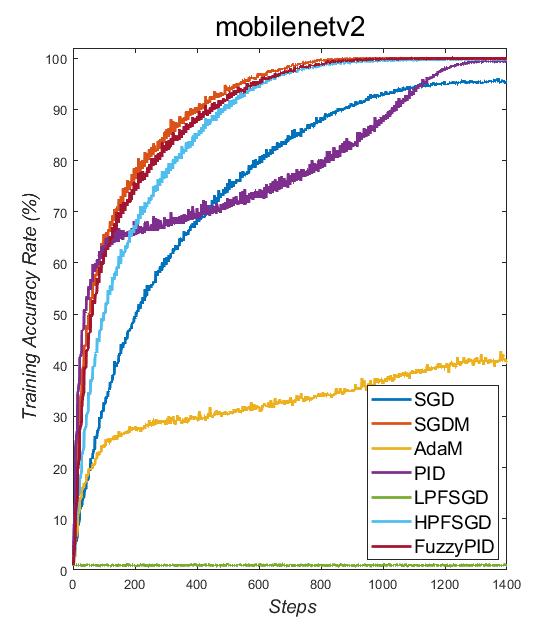} 
        % \label{fig:mean and std of net14} 
    \end{subfigure} 
    \hfill 
    \begin{subfigure}[b]{0.21\textwidth}
        % \caption{$G_{a}$ to $G_{b}$ on 1 epochs} 
        \centering 
        \includegraphics[scale=0.17]{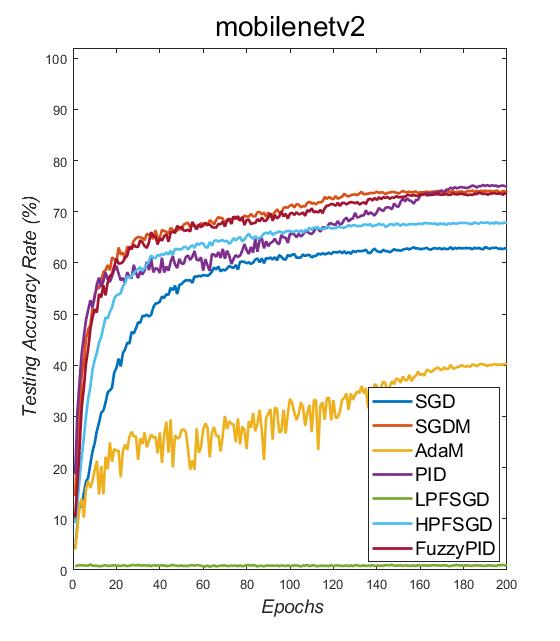} 
        % \label{fig:mean and std of net14} 
    \end{subfigure} 
    \vskip 
    \baselineskip 
    \begin{subfigure}[b]{0.21\textwidth}
        % \caption{$G_{a}$ to $G_{b}$ on 1 epochs} 
        \centering 
        \includegraphics[scale=0.17]{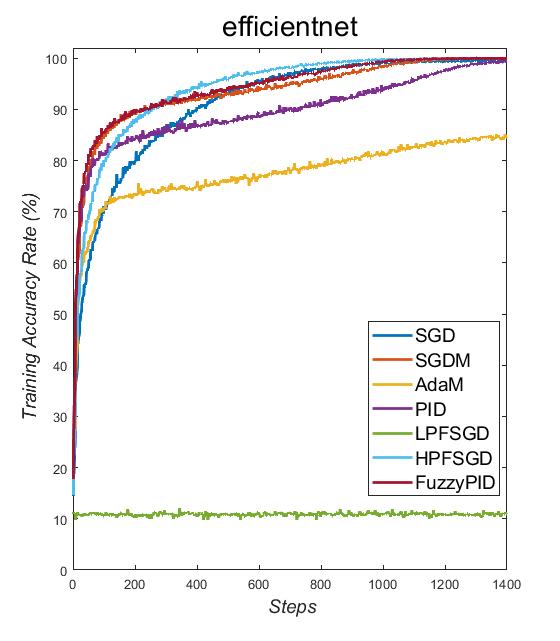} 
        % \label{fig:mean and std of net14} 
    \end{subfigure} 
    \hfill 
    \begin{subfigure}[b]{0.21\textwidth}
        % \caption{$G_{a}$ to $G_{b}$ on 1 epochs} 
        \centering 
        \includegraphics[scale=0.17]{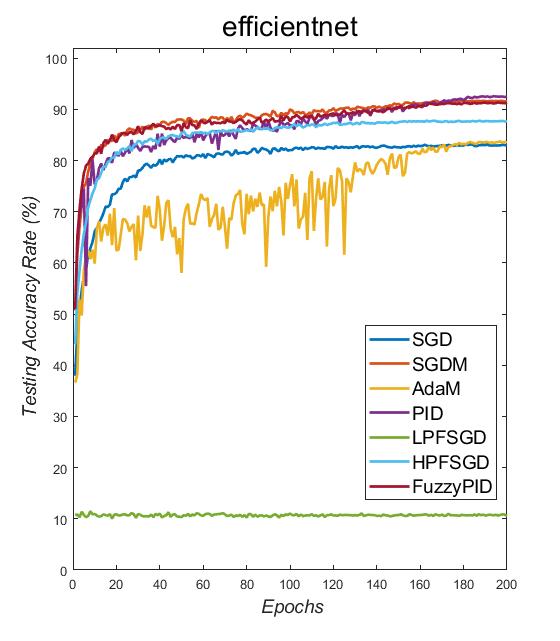} 
        % \label{fig:mean and std of net14} 
    \end{subfigure} 
    \hfill 
    \begin{subfigure}[b]{0.21\textwidth}
        % \caption{$G_{a}$ to $G_{b}$ on 1 epochs} 
        \centering 
        \includegraphics[scale=0.17]{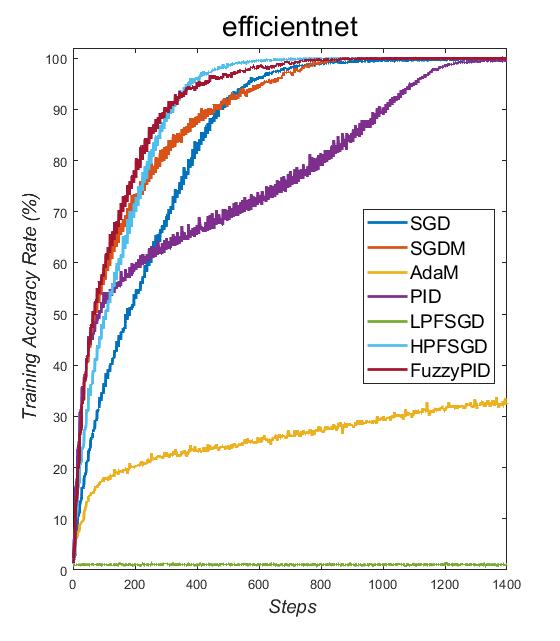} 
        % \label{fig:mean and std of net14} 
    \end{subfigure} 
    \hfill 
    \begin{subfigure}[b]{0.21\textwidth}
        % \caption{$G_{a}$ to $G_{b}$ on 1 epochs} 
        \centering 
        \includegraphics[scale=0.17]{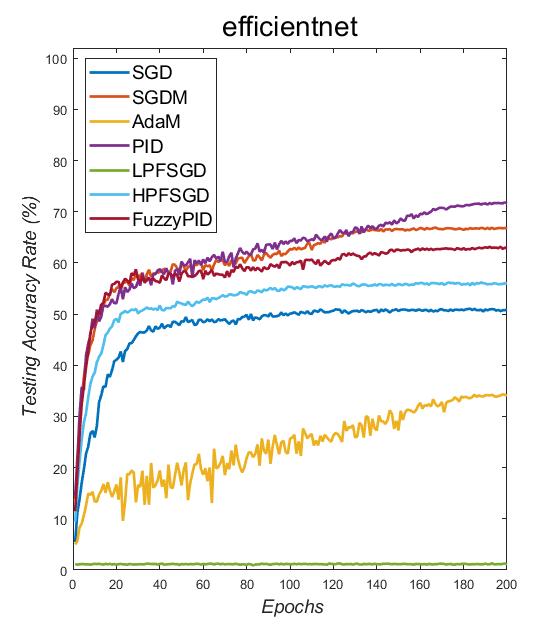} 
        % \label{fig:mean and std of net14} 
    \end{subfigure} 
    \caption{The training and testing curves of SOTA models on CIFAR10 and CIFAR100 datasets, and from the top to the bottom respectively is VGG19, ResNet50, DenseNet121, MobileNetV2 and EfficientNet on corresponding optimisers: SGD, SGDM, AdaM, PID, LPF-SGD, HPF-SGD and FuzzyPID.} 
    \label{Figure14} 
\end{figure}

\begin{figure}[h]
    \centering 
    \begin{subfigure}[b]{0.45\textwidth}
        \caption{Training Accuracy on TinyImageNet200.} 
        \centering 
        \includegraphics[scale=0.28]{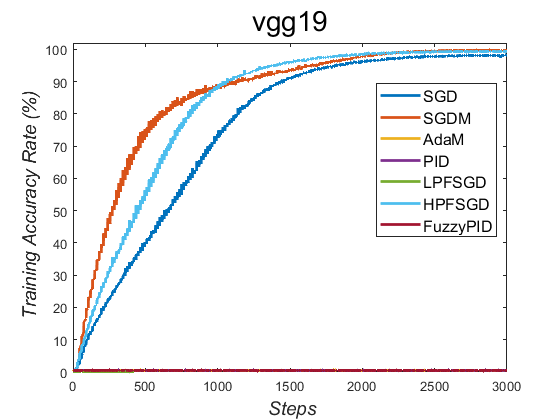} 
        % \label{fig:mean and std of net14} 
    \end{subfigure} 
    \hfill 
    \begin{subfigure}[b]{0.45\textwidth}
        \caption{Testing Accuracy on TinyImageNet200.} 
        \centering 
        \includegraphics[scale=0.28]{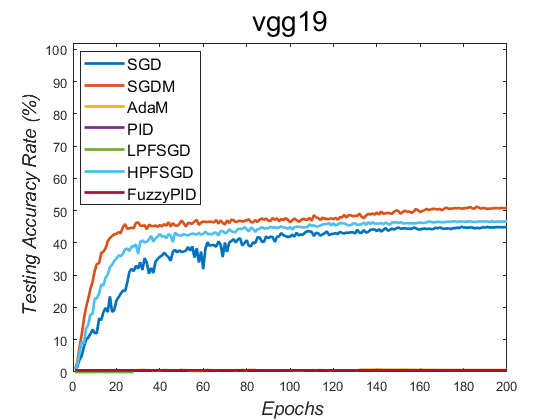} 
        % \label{fig:mean and std of net14} 
    \end{subfigure} 
    \vskip 
    \baselineskip  
    \begin{subfigure}[b]{0.45\textwidth}
        % \caption{Training Accuracy on CIFAR100.} 
        \centering 
        \includegraphics[scale=0.28]{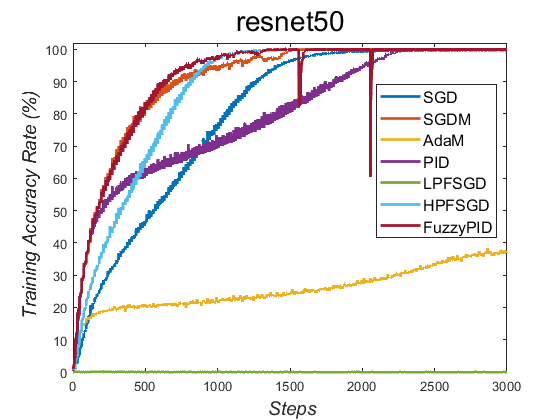} 
        % \label{fig:mean and std of net14} 
    \end{subfigure} 
    \hfill 
    \begin{subfigure}[b]{0.45\textwidth}
        % \caption{Testing Accuracy on CIFAR100.} 
        \centering 
        \includegraphics[scale=0.28]{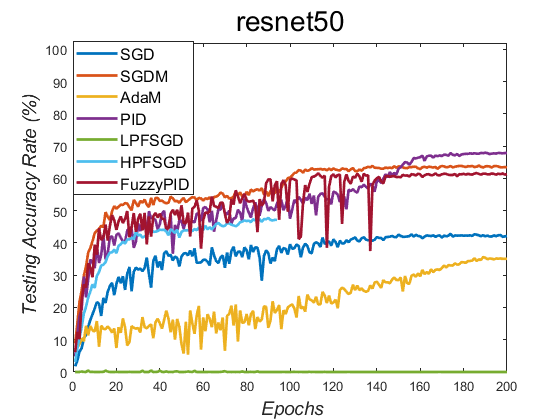} 
        % \label{fig:mean and std of net14} 
    \end{subfigure} 
    \vskip 
    \baselineskip  
    \begin{subfigure}[b]{0.45\textwidth}
        % \caption{$G_{a}$ to $G_{b}$ on 1 epochs} 
        \centering 
        \includegraphics[scale=0.28]{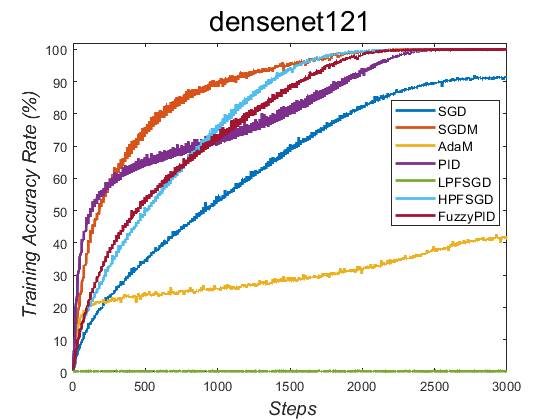} 
        % \label{fig:mean and std of net14} 
    \end{subfigure} 
    \hfill 
    \begin{subfigure}[b]{0.45\textwidth}
        % \caption{$G_{a}$ to $G_{b}$ on 1 epochs} 
        \centering 
        \includegraphics[scale=0.28]{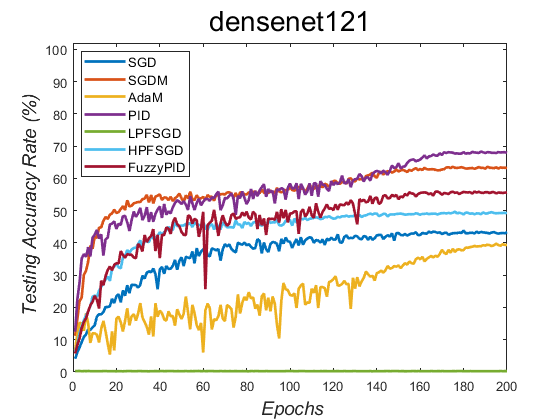} 
        % \label{fig:mean and std of net14} 
    \end{subfigure} 
    \vskip 
    \baselineskip  
    \begin{subfigure}[b]{0.45\textwidth}
        % \caption{$G_{a}$ to $G_{b}$ on 1 epochs} 
        \centering 
        \includegraphics[scale=0.28]{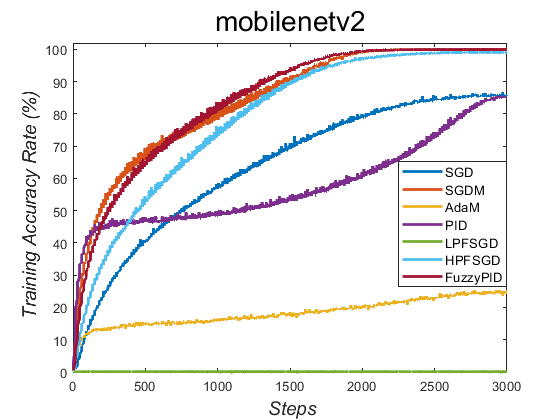} 
        % \label{fig:mean and std of net14} 
    \end{subfigure} 
    \hfill
    \begin{subfigure}[b]{0.45\textwidth}
        % \caption{Training Accuracy on CIFAR10.} 
        \centering 
        \includegraphics[scale=0.28]{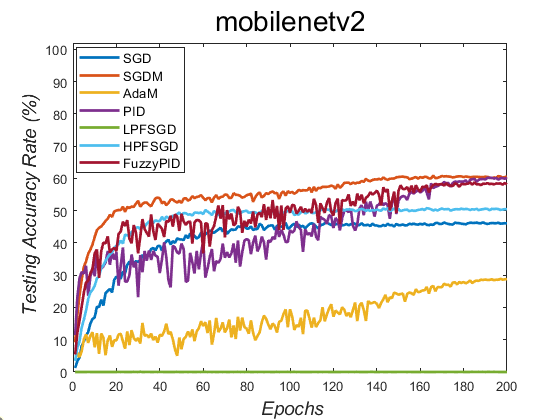} 
        % \label{fig:mean and std of net14} 
    \end{subfigure} 
    \vskip 
    \baselineskip  
    \begin{subfigure}[b]{0.45\textwidth}
        % \caption{Testing Accuracy on CIFAR10.} 
        \centering 
        \includegraphics[scale=0.28]{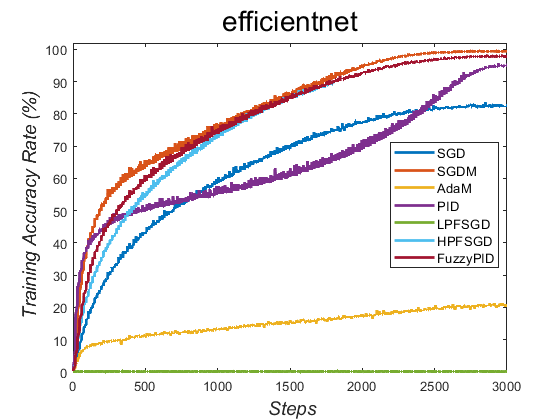} 
        % \label{fig:mean and std of net14} 
    \end{subfigure} 
    \hfill 
    \begin{subfigure}[b]{0.45\textwidth}
        % \caption{Training Accuracy on CIFAR100.} 
        \centering 
        \includegraphics[scale=0.28]{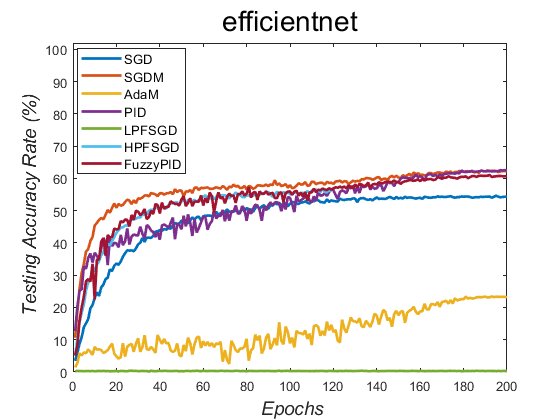} 
        % \label{fig:mean and std of net14} 
    \end{subfigure} 
    \caption{The training and testing curves of SOTA models (e.g., VGG19, ResNet50, DenseNet121, MobileNetV2 and EfficientNet) on TinyImageNet on corresponding optimisers: SGD, SGDM, AdaM, PID, LPF-SGD, HPF-SGD and FuzzyPID.} 
    \label{Figure15} 
\end{figure}

\end{document}